\documentclass[10pt,twocolumn,letterpaper]{article}

\usepackage{cvpr}              % To produce the CAMERA-READY version
\usepackage[table]{xcolor} % 색상 사용 (HTML 코드 지원, table 옵션)
\usepackage{colortbl}     % \cellcolor (셀 배경색)
\usepackage{adjustbox}    % \begin{adjustbox} (표 크기 조절)
\usepackage{multirow}     % \multirow (세로 셀 병합)
\usepackage{amsmath}      % \text{...} (수학 모드 내 텍스트)
\usepackage{float}
\usepackage{threeparttable} % 표 하단 텍스트 삽입 목적
\usepackage{booktabs}
\usepackage{soul}
\usepackage{setspace}
\usepackage{amsmath}
\usepackage{amsfonts}
\usepackage{amssymb}

\definecolor{MyRed}{HTML}{FFA0A0}
\definecolor{MyOrange}{HTML}{FFDBB0}
\definecolor{MyYellow}{HTML}{FFFFC7}

\definecolor{cvprblue}{rgb}{0.21,0.49,0.74}
\usepackage[pagebackref,breaklinks,colorlinks,allcolors=cvprblue]{hyperref}

\def\paperID{20641} % *** Enter the Paper ID here
\def\confName{CVPR}
\def\confYear{2026}

\title{PR-IQA: Partial-Reference Image Quality Assessment for Diffusion-Based Novel View Synthesis}

\author{
Inseong Choi$^{1*}$ \quad
Siwoo Lee$^{1*}$ \quad
Seung-Hun Nam$^{2\dagger}$ \quad
Soohwan Song$^{1\dagger}$\\
$^{1}$Dongguk University \quad
$^{2}$NAVER WEBTOON AI
}

\begin{document}
\maketitle
\begingroup
\renewcommand\thefootnote{}
\footnotetext{* Equal contribution. $\dagger$ Corresponding authors.}
\endgroup

\begin{abstract}
Diffusion models are promising for sparse-view novel view synthesis (NVS), as they can generate pseudo-ground-truth views to aid 3D reconstruction pipelines like 3D Gaussian Splatting (3DGS). However, these synthesized images often contain photometric and geometric inconsistencies, and their direct use for supervision can impair reconstruction. To address this, we propose Partial-Reference Image Quality Assessment (PR-IQA), a framework that evaluates diffusion-generated views using reference images from different poses, eliminating the need for ground truth. PR-IQA first computes a geometrically consistent partial quality map in overlapping regions. It then performs quality completion to inpaint this partial map into a dense, full-image map. This completion is achieved via a cross-attention mechanism that incorporates reference-view context, ensuring cross-view consistency and enabling thorough quality assessment. When integrated into a diffusion-augmented 3DGS pipeline, PR-IQA restricts supervision to high-confidence regions identified by its quality maps. Experiments demonstrate that PR-IQA outperforms existing IQA methods, achieving full-reference-level accuracy without ground-truth supervision. Thus, our quality-aware 3DGS approach more effectively filters inconsistencies, producing superior 3D reconstructions and NVS results.
\end{abstract}    
\begin{figure*}
    \centering
    \includegraphics[width=1.0\linewidth]{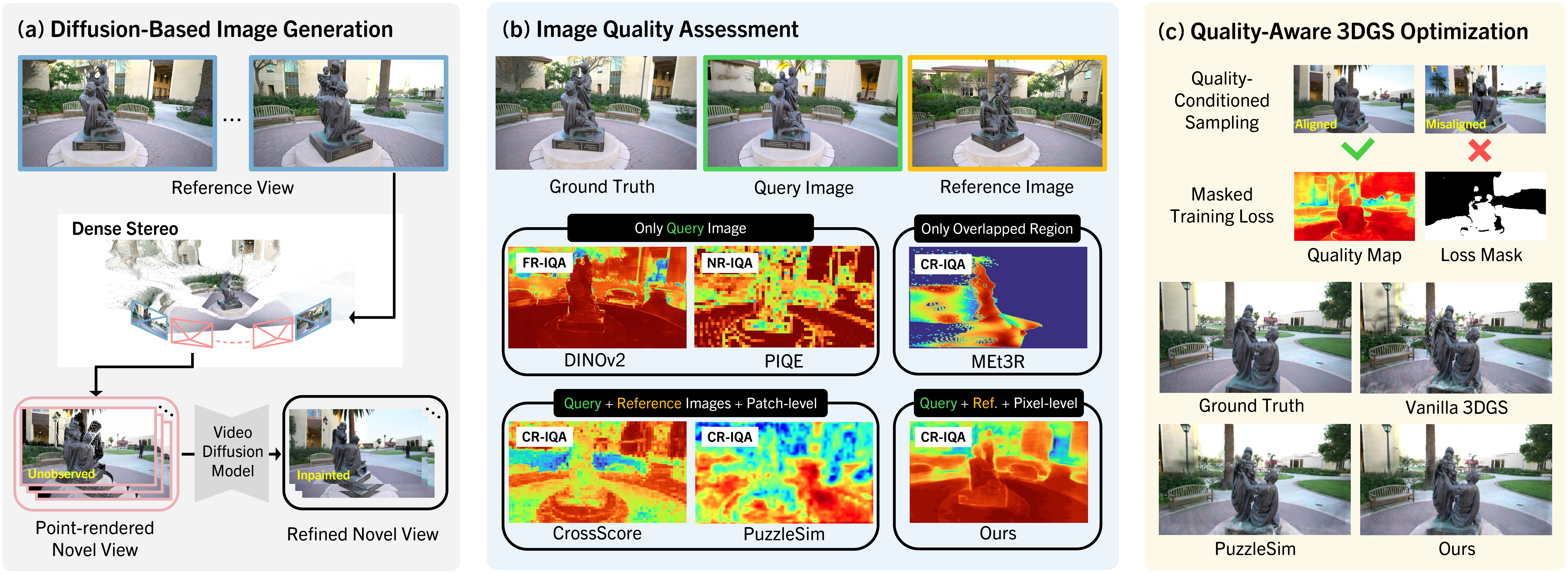}
    \caption{Overview of the proposed PR-IQA and quality-aware 3DGS. (a) Diffusion models generate novel views (pseudo-GTs) from sparse inputs, which often contain photometric or geometric artifacts. (b) We propose PR-IQA, a cross-reference method predicting a dense, pixel-level quality map from unaligned references. It produces a complete map correlating closely with FR-IQA metrics (e.g., DINOv2 feature-similarity map) without requiring a GT. (c) This quality map enables a dual-filtering strategy (image selection and pixel masking) for 3DGS training, reducing reconstruction errors and improving fidelity.}
    \label{F1_Main}
    \vspace{-2mm}
\end{figure*}

\section{Introduction}
Diffusion models~\cite{song2020denoising, ho2020denoising, rombach2022high} have become central to image-based 3D reconstruction because of their strong image-synthesis capabilities. In sparse-view settings~\cite{niemeyer2022regnerf, yu2024viewcraftertamingvideodiffusion} with limited input views, they complement conventional pipelines such as Neural Radiance Fields (NeRF)~\cite{mildenhall2021nerf} and 3D Gaussian Splatting (3DGS)~\cite{kerbl20233d} by generating novel views that act as pseudo-ground-truth views. These synthesized views densify supervision, fill coverage gaps in undersampled regions, thereby improving optimization and novel view synthesis (NVS) quality~\cite{yu2024viewcraftertamingvideodiffusion, cai2024dust, bao2025free360}. However, diffusion-generated images can exhibit photometric artifacts or geometric inconsistencies; training on them without proper quality assessment risks amplifying errors and distorting the reconstructed 3D geometry. This has led to a growing interest in \textit{Image Quality Assessment} (IQA) methods tailored to generated views. 

IQA methods are commonly grouped into \textit{full-reference} (FR) and \textit{no-reference} (NR) categories, depending on whether a reference image is available. FR metrics, such as PSNR, SSIM~\cite{wang2004image}, and LPIPS~\cite{Zhang_2018_CVPR}, can achieve high accuracy by comparing the query image against a pixel-aligned ground-truth (GT) view. However, this reliance on a GT limits their applicability to tasks like NVS and 3D reconstruction, where such GTs are often unavailable. Conversely, NR methods~\cite{mittal2012no, venkatanath2015blind, zhang2023perceptual, ying2020patches} operate without any reference image, offering greater flexibility, but they often struggle to detect the subtle artifacts and geometric inconsistencies specific to diffusion-generated images. To bridge this gap, \textit{cross-reference} (CR) evaluation~\cite{wang2024crossscore, hermann2025puzzle, asim2025met3r} has recently emerged. This method leverages multiple unaligned reference views from the same scene, combining camera geometry with photometric similarity to generate perceptually aligned quality maps without requiring GT supervision.

CR-IQA primarily relies on two strategies: \textit{patch-based similarity} and \textit{multi-view consistency}. Patch-based methods, including CrossScore~\cite{wang2024crossscore} and Puzzle Similarity~\cite{hermann2025puzzle}, evaluate photometric consistency by comparing image patches across views, making them effective at detecting local visual artifacts. However, because they depend on simple measures such as SSIM or generic CNN features, they cannot capture high-level semantic information and provide only a limited assessment of geometric alignment. In contrast, multi-view consistency methods~\cite{asim2025met3r} explicitly evaluate both geometric and photometric coherence. For example, MEt3R~\cite{asim2025met3r} warps 3D structure between views to establish correspondences and then measures feature similarity within aligned regions. This yields reliable quality estimates in observed areas but cannot assess unobserved regions, leaving a blind spot in evaluation. 

To overcome these limitations, we propose \textit{Partial-Reference} IQA (PR-IQA), a novel CR-IQA method that integrates the strengths of both patch-similarity and multi-view consistency approaches. PR-IQA operates in two stages: (i) partial quality estimation in mutually observable regions, and (ii) quality completion for non-overlapping regions using partial references. First, we warp 3D data to align query and reference views, identifying mutually visible pixels and computing a partial quality map through feature similarity in these aligned regions. Second, we formulate quality evaluation of unobserved regions as a quality completion problem. Unlike traditional image completion~\cite{criminisi2004region, pathak2016context} that predicts pixel values from local context, our approach infers quality scores for unseen areas using the partial map as guidance. This effectively extrapolates quality estimates across the entire image, mitigating blind spots and enabling more comprehensive quality assessment. 

Our quality completion network uses a novel three-stream encoder-decoder architecture that processes the query image, reference image, and partial quality map. Its core is a reference-conditioned cross-attention mechanism, injecting features from the reference encoder into the query and partial map encoders. This architecture enforces explicit view alignment and integrates cross-view evidence at all scales. Furthermore, each encoder uses a dual-gated attention block, decoupling channel and spatial attention to promote effective quality propagation into non-overlapping regions. Consequently, our method predicts quality maps with accuracy comparable to FR-IQA metrics, despite operating without GT supervision.

We also demonstrate the practical utility of PR-IQA by integrating it into a sparse-view 3DGS pipeline. This integration employs a dual-filtering strategy: (i) at the image level, we use PR-IQA to score generated pseudo-GT candidates and select the best one; (ii) at the pixel level, its dense quality map creates a binary confidence mask. This mask restricts the 3DGS optimization loss to only high-confidence regions, filtering out artifacts and inconsistencies. This quality-aware approach ensures the 3DGS model trains on the most trustworthy regions of generated views, significantly improving reconstruction fidelity. As shown in Fig.~\ref{F1_Main}, our pipeline effectively filters inconsistent content while preserving geometric fidelity, yielding accurate 3D reconstructions and high-fidelity NVS. 

The main contributions are summarized as follows:
\begin{itemize}
    \item We propose a novel CR-IQA method, PR-IQA, that reformulates quality estimation as a quality completion using a geometrically consistent partial map.
    \item We introduce a reference-conditioned quality completion network that leverages cross-attention to align views, achieving FR-level accuracy without GT supervision.
    \item We develop a quality-aware 3DGS training pipeline that leverages high-confidence regions identified by PR-IQA.
    \item We construct a new CR-IQA dataset using diffusion-generated variants of standard benchmark images, and we release both the dataset and source code. \footnote{\url{https://github.com/Kakaomacao/PR-IQA}\label{fn:github}}
\end{itemize}

\section{Related Work}
\subsection{Diffusion-based Sparse Novel View Synthesis}
Diffusion models~\cite{song2020denoising, ho2020denoising, rombach2022high} have achieved state-of-the-art performance in image, video, and 3D synthesis. They are increasingly applied to sparse-view NVS, generating new viewpoints from limited inputs~\cite{yu2024viewcraftertamingvideodiffusion}. Multi-view diffusion frameworks can infer 3D structure to produce cross-view-consistent renderings~\cite{watson2022novel, liu2023zero1to3zeroshotimage3d}, which are then fused by downstream pipelines like NeRFs~\cite{wynn2023diffusionerf, liu2024deceptivenerf3dgsdiffusiongeneratedpseudoobservationshighquality} or textured meshes~\cite{tang2024mvdiffusiondensehighresolutionmultiview,tang2023mvdiffusionenablingholisticmultiview, Liu_2024}.

Diffusion models are also used to generate pseudo-GT views for 3DGS to improve coverage in sparse-view settings. ViewCrafter~\cite{yu2024viewcraftertamingvideodiffusion} and its variants~\cite{cai2024dust, bao2025free360, yu2025wonderworld} handle extremely sparse inputs using auxiliary priors (e.g., layered scene representations~\cite{bao2025free360}, inpainting~\cite{cai2024dust}) to bridge gaps and enforce consistency. However, these synthesized views often contain geometric artifacts or inconsistent textures. Naively using such views without quality filtering can degrade the reconstruction. 

Only a few studies have attempted to mitigate this problem. Wang et al.~\cite{wang2025active} proposed image-level quality scores for selection, but this still allows artifact-prone regions to be used in optimization. Another approach~\cite{bao2025free360} proposed measuring pixel-wise uncertainty for the generated images and selectively applying this to the 3DGS training. This method, however, defines uncertainty simply as the pixel-wise variance across multiple generated images for a single view. This definition is highly dependent on the output distribution of the specific generative model and often leads to inaccurate uncertainty predictions.

 In contrast, our work proposes a network that directly predicts image quality. We leverage these predictions for both robust image selection and pixel-wise adaptive 3DGS training. This dual-filtering approach ensures that only the most accurate regions of the generated images contribute to the reconstruction, thereby significantly improving 3DGS modeling performance.

\begin{figure*}
    \centering
    \includegraphics[width=0.98\linewidth]{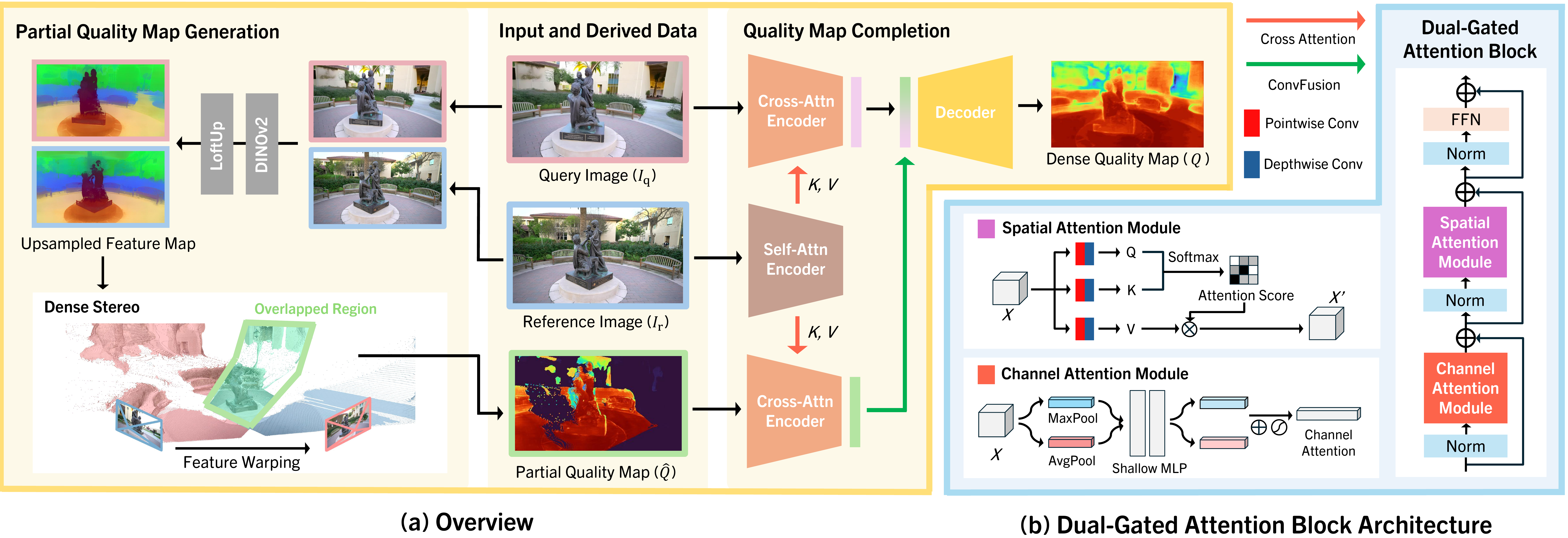}
    \vspace{-2mm}
    \caption{(a) Overview of the PR-IQA pipeline. The framework operates in two stages. First, we warp DINOv2 features from the reference $I_r$ to the query $I_q$ view via dense stereo, generating a partial quality map ($\hat{Q}$) for overlapping regions. Next, a three-stream (query, reference, partial map) encoder-decoder predicts the full quality map $Q$. (b) Architecture of the Dual-Gated Attention Block. The block sequentially applies two attention mechanisms: a Channel Attention Module (using max/avg pooling and MLP) recalibrates channels, and a Spatial Attention Module (using Q, K, V projections and softmax) provides spatial refinement. The block integrates both with normalization, residual connections ($\oplus$), and an FFN. Each encoder and decoder is composed of this block.}
    \label{F2_Overview}
    \vspace{-2mm}
\end{figure*}

\subsection{Image Quality Assessment}
Full-reference (FR-IQA) metrics assess image quality by comparing a query image against a pose-aligned GT. Established metrics like PSNR and SSIM~\cite{wang2004image}, along with learned measures such as LPIPS~\cite{Zhang_2018_CVPR}, operate in deep feature spaces to better reflect human perception. However, their strict requirement for a pixel-aligned GT restricts their use in practical applications like NVS, where GTs are inherently unavailable.

No-reference (NR-IQA) methods bypass the need for a reference, estimating quality directly from the query image. Classical approaches (BRISQUE~\cite{mittal2012no}, PIQE~\cite{venkatanath2015blind}) use handcrafted features, while modern learning-based methods (PAL4VST~\cite{zhang2023perceptual}, PaQ-2-PiQ~\cite{ying2020patches}) leverage deep networks. Lacking a reference makes defining an objective quality standard difficult. Consequently, NR-IQA metrics are primarily designed to detect low-level artifacts, and thus are ill-suited for assessing the high-level geometric and multi-view consistency required for NVS.

To bridge this gap, cross-reference (CR-IQA) methods~\cite{wang2024crossscore, hermann2025puzzle, asim2025met3r} have emerged, which utilize unregistered reference images from the same scene. Pioneering works include CrossScore~\cite{wang2024crossscore}, which estimates SSIM maps via cross-attention, and Puzzle Similarity~\cite{hermann2025puzzle}, which employs patch-level cosine similarity. While innovative, these methods are often limited to simple patch-level comparisons and lack high-level semantic understanding. More critically, they ignore multi-view geometry, a cornerstone of reliable 3D reconstruction. MEt3R~\cite{asim2025met3r} addresses this by explicitly quantifying geometric alignment using DINO features~\cite{caron2021emerging}, but its analysis is confined to overlapping regions.

In contrast, our method unifies patch-similarity and multi-view consistency within a novel quality completion framework. Our approach first computes a geometrically aligned, reliable quality map only in overlapping regions. It then propagates trusted quality signals to non-overlapping areas via a novel cross-attention network, yielding a complete and geometrically-aware quality assessment.

\section{Partial-Reference IQA}
\subsection{Preliminaries}
\label{3.1_preliminaries}
Given a query image $I_q \in \mathbb{R}^{H \times W \times 3}$ and a reference image $I_r \in \mathbb{R}^{H \times W \times 3}$ captured from different viewpoints of the same scene, our task, CR-IQA, aims to predict a dense quality map $Q \in \left[0, 1\right]^{H \times W}$ for $I_q$. In the context of sparse NVS, $I_q$ is a pose-conditioned diffusion rendering, whereas $I_r$ is a real image from another pose. Unlike FR-IQA~\cite{wang2004image, Zhang_2018_CVPR}, we do not assume access to a pixel-aligned GT image $I_q^\ast$.

Let $f$ denote an FR quality function $f(I_q, I_q^\ast) \to Q_{\text{FR}}$, where $Q_{\text{FR}} \in [0, 1]^{H \times W}$ is the per-pixel quality map derived by comparing $I_q$ to its GT $I_q^\ast$. Our goal is to learn a cross-reference function $g_\Phi(I_q, I_r) \to Q$. This function is trained to approximate the FR-like quality map ($Q \approx Q_{\text{FR}}$) using only an unregistered reference image $I_r$ in place of $I_q^\ast$. Any established FR-IQA metric can serve as the target for the FR quality function $f$. We adopt DINOv2~\cite{oquab2023dinov2} feature similarity and SSIM, as they are strong-performing metrics for sparse NVS (see Appendix). The DINOv2 similarity is computed as the pixel-wise cosine similarity between DINOv2 feature maps extracted from both images.

\subsection{Overview}
\label{3.2_overview}
Fig.~\ref{F2_Overview}(a) illustrates our proposed CR-IQA pipeline, namely PR-IQA, which operates in two main stages: (i) partial quality map generation and (ii) dense quality map completion. In the first stage, inspired by MEt3R~\cite{asim2025met3r}, we leverage the principle that geometric overlap provides locally reliable cross-view consistency. We identify overlapping regions between $I_q$ and $I_r$ and compute a partial quality map $\hat{Q}$ exclusively for these regions. In the second stage, we formulate quality estimation for non-overlapping regions as a \textit{quality map completion} problem. Our PR-IQA network takes the partial quality map $\hat{Q}$  as input, along with the query and reference images $I_q$ and $I_r$. The network then predicts a complete quality map $Q$ over the entire image domain. This allows the network to propagate reliable quality signals from validated regions to the rest of the image, using the reference image for context guidance.

\subsection{Partial Quality Map Generation}
\label{3.3_partial}
We construct a partial quality map $\hat{Q}$ that measures feature-space consistency between a query image $I_q$ and a reference image $I_r$ only at geometry-consistent pixels. Concretely, we first obtain dense, pixel-aligned 3D point maps using visual geometry grounded transformer (VGGT)~\cite{Wang_2025_CVPR} to establish geometric correspondences. For feature comparison, we extract DINOv2~\cite{oquab2023dinov2} features ($F_q^{\text{DINO}}$, $F_r^{\text{DINO}}$) and upsample them to high-resolution using LoftUp~\cite{huang2025loftuplearningcoordinatebasedfeature}. Using the VGGT point maps, we warp the reference features $F_r^{\text{DINO}}$ into the query view via unprojection and reprojection, yielding the warped features $F_{r \to q}^{\text{DINO}}$.

The partial quality at pixel $i$ is then computed as the cosine similarity between the query features $F_q^{\text{DINO}}(i)$ and the warped reference features $F_{r \to q}^{\text{DINO}}(i)$:
\begin{equation}
\hat{Q}(i) = \text{CosSim}\left(F_q^{\text{DINO}}(i), F_{r \to q}^{\text{DINO}}(i)\right),
\label{eq:1}
\end{equation}
where $\text{CosSim}(\mathbf{u}, \mathbf{v}) = \frac{1}{2} \left( \frac{\mathbf{u} \cdot \mathbf{v}}{\|\mathbf{u}\| \|\mathbf{v}\|} + 1 \right)$ denotes the cosine similarity normalized to the range $[0, 1]$.

\subsection{Quality Map Completion}
\label{3.4_dense-quality}
Our goal is to predict a dense quality map $Q$ for $I_q$, given $I_r$ and a partial quality map $\hat{Q}$. We frame this as a cross-view quality completion task, which propagates reliable scores from overlapping regions to unobserved areas, guided by cross-view geometric and semantic consistency.

Our network architecture is composed of three encoders, $\mathrm{Enc}_{\mathrm{self}}^r$, $\mathrm{Enc}_{\mathrm{cross}}^q$, and $\mathrm{Enc}_{\mathrm{cross}}^p$, and a single decoder, $\mathrm{Dec}$. The self-attention encoder, $\mathrm{Enc}_{\mathrm{self}}^r$, processes $I_r$ to extract its features. These features are then utilized as context by the two cross-attention encoders: $\mathrm{Enc}_{\mathrm{cross}}^q$, which processes the query image $I_q$, and $\mathrm{Enc}_{\mathrm{cross}}^p$, which processes the partial quality map $\hat{Q}$. The decoder, $\mathrm{Dec}$, then synthesizes the information from the encoders to produce the final full quality map $Q$.

All encoders are multi-scale pyramids featuring three downsampling stages. As illustrated in Fig.~\ref{F2_Overview}(b), each stage employs a \textit{dual-gated attention block} derived from the CBAM model~\cite{woo2018cbam}, which utilizes a sequential attention mechanism consisting of channel attention, spatial attention, and a feed-forward network (FFN). This sequential design is highly advantageous for quality map completion, as it allows the network to decouple what features are relevant (via channel attention) from where they should be propagated to fill non-overlapping regions (via spatial attention). To stabilize matching across pose changes, we inject 2D positional encodings at every scale. The decoder mirrors this multi-scale attention pattern, progressively upsampling the fused features back to full resolution to produce the quality map $Q$.

For each stage $s$, the self-attention block $\mathrm{Enc}_{\mathrm{self}}^{r,s}$ processes the reference features, where $s \in \{0,1,2,3\}$.
The cross-attention blocks $\mathrm{Enc}_{\mathrm{cross}}^{q,s}$ and $\mathrm{Enc}_{\mathrm{cross}}^{p,s}$ are reference-conditioned: they replace the self-attention layer with a cross-attention module that takes the branch’s own features as queries while using the same-stage reference features as keys and values. This design explicitly enforces alignment to $I_r$ and injects cross-view evidence at every scale, unlike prior designs that rely solely on self-attention~\cite{vaswani2017attention}.

Let $F_r^s$, $F_q^s$, and $F_p^s$ denote the reference, query, and partial-map features at stage $s$. We initialize $F_r^0 = I_r$, $F_q^0 = I_q$, and $F_p^0 = \hat{Q}$. The computation at each stage proceeds as:
\begin{equation}
\begin{aligned}
F_r^s &= \mathrm{Enc}_{\mathrm{self}}^{r,s}(F_r^{s-1}), \\
\hat{F}_q^s &= \mathrm{Enc}_{\mathrm{cross}}^{q,s}(F_q^{s-1}; F_r^s), \\
F_p^s &= \mathrm{Enc}_{\mathrm{cross}}^{p,s}(F_p^{s-1}; F_r^s).
\end{aligned}
\label{eq:2}
\end{equation}

After each stage, we fuse the query and partial streams to update the partial representation:
\begin{equation}
F_q^s = \text{ConvFuse}(\hat{F}_q^s, F_p^s),
\label{eq:3}
\end{equation}
where $\text{ConvFuse}$ denotes channel-wise concatenation followed by a channel-mixing convolution. This fusion anchors quality propagation to geometry-validated regions from $\hat{Q}$ and steers subsequent updates toward a cross-view-consistent solution.

The final fused representation $F_q^3$ is decoded to a full-resolution map:
\begin{equation}
Q = \text{Dec}(F_q^3).
\label{eq:4}
\end{equation}

The combination of reference-conditioned cross-attention and overlap-guided fusion enforces strict geometric alignment across views, allowing the network to propagate reliable scores from overlapping to unseen regions while reducing ghosting and view-mismatch artifacts to yield a complete, perceptually coherent quality map.

\begin{table*}
\centering
\caption{Quantitative comparisons of predicted quality maps from IQA methods against GT quality maps (PLCC $\uparrow$, SRCC $\uparrow$). Red, orange, and yellow cells denote the 1st, 2nd, and 3rd best methods per column (with rankings computed excluding FR settings$^{\dagger}$), and gray cells indicate identity cases where the IQA prediction matches the GT quality map.}
\begin{adjustbox}{max width=\linewidth}
\begin{tabular}{llcccccccccccc}
\hline
                           &                         & \multicolumn{4}{c}{Mip-NeRF 360}                                           & \multicolumn{4}{c}{Tanks and Temples}                                      & \multicolumn{4}{c}{RealEstate10K}                                          \\ \cline{3-14}
IQA Type                   & IQA Method                & \multicolumn{2}{c}{DINOv2}   & \multicolumn{2}{c}{SSIM}     & \multicolumn{2}{c}{DINOv2}   & \multicolumn{2}{c}{SSIM}     & \multicolumn{2}{c}{DINOv2}   & \multicolumn{2}{c}{SSIM}     \\ \cline{3-14}
                           &                           & PLCC          & SRCC          & PLCC          & SRCC          & PLCC          & SRCC          & PLCC          & SRCC          & PLCC          & SRCC          & PLCC          & SRCC          \\ \hline
                           & PSNR$^\dagger$                      & 0.407         & 0.338         & 0.517         & 0.487         & 0.405         & 0.367         & 0.486         & 0.487         & 0.248         & 0.241         & 0.392         & 0.386         \\
                           & SSIM$^\dagger$                      & 0.409         & 0.386         & \cellcolor[HTML]{C0C0C0} 1.000         & \cellcolor[HTML]{C0C0C0} 1.000        & 0.429         & 0.423         & \cellcolor[HTML]{C0C0C0} 1.000        & \cellcolor[HTML]{C0C0C0} 1.000         & 0.400         & 0.444         & \cellcolor[HTML]{C0C0C0} 1.000         & \cellcolor[HTML]{C0C0C0} 1.000         \\
                           & LPIPS$^\dagger$                     & 0.557         & 0.472         & 0.565         & 0.554         & 0.591         & 0.590         & 0.598         & 0.595         & 0.489         & 0.516         & 0.452         & 0.460         \\
\multirow{-4}{*}{FR-IQA} & DINOv2$^\dagger$                    & \cellcolor[HTML]{C0C0C0} 1.000         & \cellcolor[HTML]{C0C0C0} 1.000         & 0.409         & 0.386         & \cellcolor[HTML]{C0C0C0} 1.000         & \cellcolor[HTML]{C0C0C0} 1.000        & 0.423         & 0.417         & \cellcolor[HTML]{C0C0C0} 1.000        & \cellcolor[HTML]{C0C0C0} 1.000        & 0.400         & 0.533         \\ \hline \hline
                           & PAL4VST                   & 0.030         & 0.031         & 0.014         & 0.014         & 0.002         & 0.001         & 0.003         & 0.004         & 0.094         & 0.088         & 0.043         & 0.048         \\
                           & PaQ-2-PiQ                 & -0.088        & -0.107        & -0.163        & -0.174        & 0.039         & 0.118         & -0.086        & -0.089        & -0.111        & -0.119        & -0.251        & -0.268        \\
\multirow{-3}{*}{NR-IQA}   & PIQE                      & 0.144         & 0.161         & -0.002        & 0.017         & 0.194         & 0.201         & 0.365         & \cellcolor[HTML]{FFFFC7}0.399         & 0.191         & 0.245         & \cellcolor[HTML]{FFDBB0}0.444         & \cellcolor[HTML]{FFDBB0}0.533         \\ \hline
                           & MEt3R* & 0.105         & 0.129         & 0.037         & 0.032         & 0.142         & 0.153         & 0.110         & 0.130         & 0.312         & 0.368         & 0.195         & 0.217         \\
                           & CrossScore                & 0.094         & 0.090         & \cellcolor[HTML]{FFDBB0}0.290         & \cellcolor[HTML]{FFDBB0}0.325         & 0.237         & 0.272         & \cellcolor[HTML]{FFDBB0}0.444         & \cellcolor[HTML]{FFDBB0}0.462         & 0.285         & 0.324         & \cellcolor[HTML]{FFFFC7}0.442         & \cellcolor[HTML]{FFFFC7}0.523         \\
                           & PuzzleSim                 & 0.304         & 0.327         & 0.128         & 0.124         & \cellcolor[HTML]{FFFFC7}0.351         & \cellcolor[HTML]{FFFFC7}0.369         & 0.348         & 0.347         & \cellcolor[HTML]{FFDBB0}0.410         & \cellcolor[HTML]{FFFFC7}0.478         & 0.384         & 0.415         \\
                           & Ours$_{\text{partial}}$* & \cellcolor[HTML]{FFDBB0}0.437         & \cellcolor[HTML]{FFDBB0}0.596         & 0.150         & 0.169         & \cellcolor[HTML]{FFDBB0}0.407         & \cellcolor[HTML]{FFDBB0}0.557         & 0.098         & 0.116         & \cellcolor[HTML]{FFFFC7}0.325         & \cellcolor[HTML]{FFDBB0}0.509         & 0.206         & 0.266         \\
                           & Ours$_{\text{DINOv2}}$      & \cellcolor[HTML]{FFA0A0}0.555         & \cellcolor[HTML]{FFA0A0}0.622         & \cellcolor[HTML]{FFFFC7}0.261         & \cellcolor[HTML]{FFFFC7}0.241         & \cellcolor[HTML]{FFA0A0}0.573         & \cellcolor[HTML]{FFA0A0}0.650         & \cellcolor[HTML]{FFFFC7}0.387         & 0.367         & \cellcolor[HTML]{FFA0A0}0.453         & \cellcolor[HTML]{FFA0A0}0.564         & 0.352         & 0.395         \\
\multirow{-6}{*}{CR-IQA}   & Ours$_{\text{SSIM}}$       & \cellcolor[HTML]{FFFFC7}0.320         & \cellcolor[HTML]{FFFFC7}0.367         & \cellcolor[HTML]{FFA0A0}0.535         & \cellcolor[HTML]{FFA0A0}0.556         & 0.309         & 0.345         & \cellcolor[HTML]{FFA0A0}0.625         & \cellcolor[HTML]{FFA0A0}0.643         & 0.278         & 0.324         & \cellcolor[HTML]{FFA0A0}0.632         & \cellcolor[HTML]{FFA0A0}0.677         \\ \hline
\end{tabular}
\end{adjustbox}
\begin{tablenotes}
\footnotesize
\item $\dagger$ Metrics require a same-pose GT image.  * Metrics are computed only over the valid overlapping region.
\end{tablenotes}
\label{T1_Quanti_QualityMap}
\vspace{-2mm}
\end{table*}

\subsection{Training Strategy}
\label{3.5_training}
We train our model to predict a quality map $Q$ that approximates a GT map $Q^\ast$. This target map is derived using a custom FR-IQA metric that requires a pixel-aligned GT image $I_q^\ast$, which is unavailable at inference. We define the target quality map $Q^\ast$ as either the DINOv2 feature-similarity map (DINOv2-SIM) or the SSIM map computed between $I_q^\ast$ and $I_q$. Consequently, we train two separate model variants, each targeting one of these two FR-IQA metrics.

The training objective combines three distinct loss components.
First, a pixel-wise $\mathcal{L}_{1}$ loss $\mathcal{L}_{1}^{\text{IQA}}$ between $Q$ and $Q^\ast$ ensures local accuracy.
Second, to align the global score distributions, we apply a Jensen-Shannon Divergence (JSD)~\cite{englesson2021generalizedjensenshannondivergenceloss} loss, $\mathcal{L_{\mathrm{JSD}}}$, between $Q$ and $Q^\ast$. Third, a Pearson Linear Correlation Coefficient (PLCC)~\cite{cao2022pkdgeneraldistillationframework} loss, $\mathcal{L_{\mathrm{PLCC}}}$, is used to enforce linear agreement between $Q$ and $Q^\ast$.

Finally, the total training loss $\mathcal{L}$ is a weighted combination of these three components:
\begin{equation}
\mathcal{L} = \lambda_{\text{IQA}} \mathcal{L}_{1}^{\text{IQA}} + \lambda_{\mathrm{JSD}} \mathcal{L_{\mathrm{JSD}}} + \lambda_{\mathrm{PLCC}} \mathcal{L_{\mathrm{PLCC}}},
\label{eq:5}
\end{equation}
with weights $\lambda_{\text{IQA}}=0.5$, $\lambda_{\mathrm{JSD}}=1.0$, and $\lambda_{\mathrm{PLCC}}=0.25$.

\section{PR-IQA-Guided 3D Gaussian Splatting}
\label{4_priqa_gs}
We integrate our PR-IQA framework into a sparse-view 3DGS pipeline to filter inconsistencies from diffusion-generated views. We adopt ViewCrafter~\cite{yu2024viewcraftertamingvideodiffusion} as our baseline for generating pseudo-GT images. Fig.~\ref{F1_Main}(c) illustrates our quality-aware approach, which involves a two-stage process: (i) robust pseudo-GT selection using PR-IQA scores, and (ii) pixel-wise adaptive 3DGS training using the high-confidence regions identified by our quality maps.

\noindent\textbf{Pseudo-GT Generation and Selection.}
Given a sparse set of $N_{\text{in}}$ input images $\mathcal{I}_{\text{in}}$ = $\{{I}_{\text{in}}^k\}_{k=1}^{N_{\text{in}}}$ with camera parameters recovered via DUSt3R~\cite{wang2024dust3rgeometric3dvision}, we first sample new viewpoints. Following ViewCrafter~\cite{yu2024viewcraftertamingvideodiffusion}, we identify two nearby inputs, $I_{r1}$ and $I_{r2}$ from $\mathcal{I}_{\text{in}}$, and sample viewpoints $v$ along the trajectory between them.

For each viewpoint $v$, the video diffusion model (VDM)~\cite{yu2024viewcraftertamingvideodiffusion} generates $N_v$ candidate images, denoted as $\mathcal{I}_v$ = $\{I_{v,n}\}_{n=1}^{N_v}$. Each candidate $I_{v,n}$ (acting as the query image $I_q$) is then evaluated by our PR-IQA model $g_\Phi$, using $I_{r1}$ and $I_{r2}$ as reference images. Let $Q_{v,n}^{r1}$ and $Q_{v,n}^{r2}$ be the predicted quality maps for $I_{v,n}$ using $I_{r1}$ and $I_{r2}$ as references, respectively. We take the pixel-wise maximum of the two quality maps to form a consolidated map $Q_{v,n}$. This optimistic aggregation strategy ensures that a region is considered high-quality if it is consistent with at least one of the reference views, which is crucial for effectively expanding the training set with pseudo-GTs. Finally, a representative image-level quality score $S_{v,n}$ is then defined as the mean of all values in $Q_{v,n}$.

For each viewpoint $v$, we select the single candidate with the highest score as the pseudo-GT image $\tilde{I}_v$ and retain its corresponding quality map $\tilde{Q}_v$:
\begin{equation}
(\tilde{I}_v, \tilde{Q}_v) = \operatornamewithlimits{argmax}_{(I_{v,n}, Q_{v,n}) \in \mathcal{I}_v} (S_{v,n}).
\label{eq:6}
\end{equation}
This selection process yields a set of high-quality pseudo-GT images, $\mathcal{I}_{\text{pseudo}} = \{\tilde{I}_v\}$, which are used to densify the sparse input views.

\noindent\textbf{Quality-Aware 3DGS Training.}
We perform pixel-wise adaptive 3DGS training using the full set of images $\mathcal{I}_{\text{train}} = \mathcal{I}_{\text{in}} \cup \mathcal{I}_{\text{pseudo}}$. We define a binary confidence mask $M_k$ for each training image $I_k \in \mathcal{I}_{\text{train}}$. For real input images ($I_k \in \mathcal{I}_{\text{in}}$), the mask $M_k$ is set to all ones, as we trust all pixels. For pseudo-GT images ($I_k \in \mathcal{I}_{\text{pseudo}}$), the mask is derived from its corresponding quality map $Q_k$ to restrict training to only high-confidence regions. This is achieved by thresholding the map at its top $\tau$-percentile. Formally, the mask for a pixel $i$ is $M_k(i) = \mathbf{1}[Q_k(i) \ge T_\tau]$, where $\mathbf{1}[\cdot]$ is the indicator function and $T_\tau = \text{percentile}(Q_k, \tau)$. We heuristically set $\tau=50$.

Let $\hat{I}_k$ be the image rendered from the 3D Gaussians at the $k$-th pose. We employ the pixel-wise adaptive $\mathcal{L}_{1}$ loss $\mathcal{L}_{1,k}^{\text{3DGS}}$, which computes the difference between $I_k$ and $\hat{I}_k$, but only for pixels within high-confidence regions specified by the mask $M_k$. The final loss is then a combination of this masked reconstruction loss
$\mathcal{L}_{1,k}^{\text{3DGS}}$ and the SSIM term~\cite{kerbl20233d}:
\begin{equation}
\begin{aligned}
\mathcal{L}_{\text{total}} = \sum_{k=1}^{|\mathcal{I}_{\text{train}}|} \left( (1 - \lambda_{\text{dssim}}) \mathcal{L}_{1,k}^{\text{3DGS}} + \lambda_{\text{dssim}} \mathcal{L}_{\text{dssim}}(\hat{I}_k, I_k) \right),
\label{eq:7}
\end{aligned}
\end{equation}
where $\lambda_{\text{dssim}} = 0.2$. This quality-aware formulation supervises the 3DGS optimization using reliable real images, while leveraging high-confidence regions from synthesized views to filter photometric and geometric inconsistencies.

\begin{figure*}
    \centering
    \includegraphics[width=0.98\linewidth]{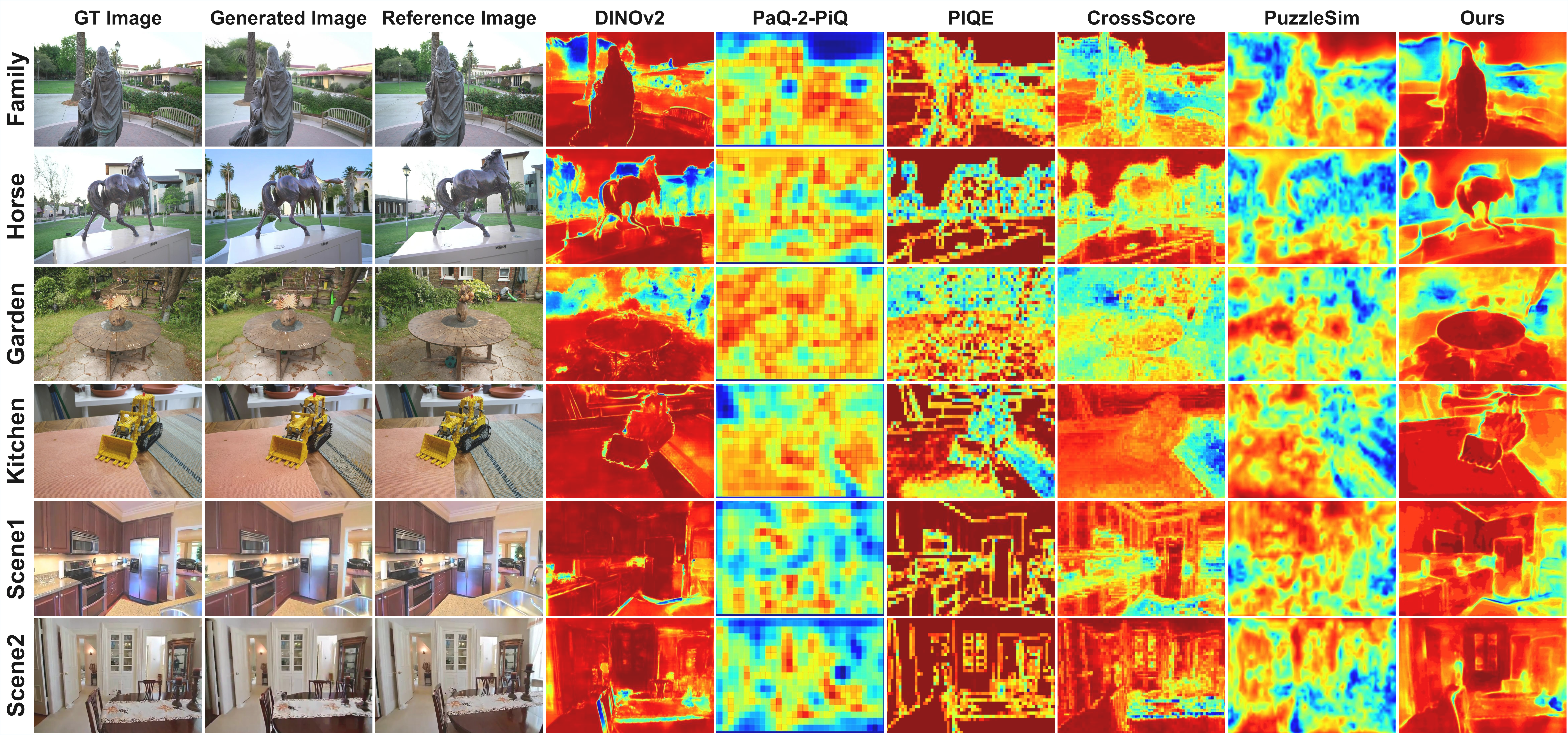}
    \caption{Qualitative comparison of estimated quality maps from IQA methods. Colors encode estimated quality, where low-quality pixels are shown in blue and high-quality pixels are shown in red. Compared to baselines, our results (``Ours") more faithfully recover object silhouettes and fine structures, closely matching the GT (DINOv2-SIM).}
    \label{F3_Quali_Score}
    \vspace{-1mm}
\end{figure*}

\section{Experiments}
\subsection{Experimental Settings}
\noindent\textbf{Dataset.} We constructed our training dataset using the MFR dataset~\cite{arnold2022map} by uniformly sampling GT frames and using a VDM~\cite{yu2024viewcraftertamingvideodiffusion} to generate three variant query images per frame, resulting in 120k training pairs. For evaluation, we used three benchmarks: Tanks and Temples~\cite{10.1145/3072959.3073599}, Mip-NeRF 360~\cite{barron2022mip}, and RealEstate10K~\cite{zhou2018stereomagnificationlearningview}. For each benchmark, we pre-generated a fixed set of query images (using the same VDM) and their corresponding GT images, which are used only for computing the ground-truth quality maps during evaluation. Our benchmark is publicly available$^{\ref{fn:github}}$; details appear in the Appendix. 

\noindent\textbf{Baselines.} We compare our method against three categories of baselines: FR-IQA, NR-IQA, and CR-IQA. All IQA methods are configured to produce pixel-wise quality maps for our alignment-based comparison. For FR-IQA, we include well-established metrics, PSNR, SSIM~\cite{wang2004image}, LPIPS~\cite{Zhang_2018_CVPR}, and DINOv2-SIM, each computed against the pose-aligned GT image. For NR-IQA, we employ PAL4VST~\cite{zhang2023perceptual}, PaQ-2-PiQ~\cite{ying2020patches}, and PIQE~\cite{venkatanath2015blind}. For CR-IQA, we evaluate CrossScore~\cite{wang2024crossscore}, PuzzleSim~\cite{hermann2025puzzle}, and MEt3R~\cite{asim2025met3r}. Our PR-IQA method, as introduced in Sect.~\ref{3.2_overview}, is evaluated in two variants: Ours$_{\text{DINOv2}}$ and Ours$_{\text{SSIM}}$, which target the DINOv2 and SSIM metrics, respectively. To assess the partial map's utility, we also include Ours$_{\text{partial}}$, a non-learned variant using the partial map directly.

\subsection{IQA Performance}
\noindent\textbf{Setup.} We evaluate competing IQA methods by measuring their alignment with GT quality maps. GT maps are generated by comparing diffusion outputs (query images) to their pose-aligned GT counterparts; we consider two variants: DINOv2-SIM and SSIM map. We quantify map alignment using the Pearson Linear Correlation Coefficient (PLCC) and the Spearman Rank Correlation Coefficient (SRCC). PLCC measures linear correlation (invariant to affine scaling), and SRCC captures rank-order correlation (invariant to monotonic transformations).

Following standard protocols, FR-IQA uses the query and its pose-aligned GT image; NR-IQA uses only the query. For CR-IQA, which measures cross-view consistency, we use the closer of the first or last frame as a reference view for each query, with the intermediate frames serving as the query views. Predicted maps are generated for these reference-query pairs, and alignment is evaluated on all intermediate frames. Since Ours$_{\text{partial}}$ and MEt3R~\cite{asim2025met3r} operate on sub-regions, their evaluation is restricted to the valid support, computing PLCC and SRCC only within these corresponding spatial regions.

\noindent\textbf{Results.} Table~\ref {T1_Quanti_QualityMap} summarizes the average PLCC and SRCC for each IQA metric. As expected, FR-IQA methods achieve the highest performance given their access to pose-aligned GT images, with LPIPS yielding the best results on both DINOv2 and SSIM targets. Conversely, NR-IQA methods demonstrate the lowest average performance. The performance of CR-IQA baselines aligns with their respective designs: CrossScore (trained on SSIM) exhibits high correlation with SSIM, while PuzzleSim (feature-based) correlates more strongly with DINOv2.

Notably, our PR-IQA variants, Ours$_{\text{DINOv2}}$ and Ours$_{\text{SSIM}}$, achieve state-of-the-art performance in their categories. As confirmed by Table~\ref{T1_Quanti_QualityMap}, Ours$_{\text{DINOv2}}$ effectively targets the feature-similarity map from a large-scale backbone. Ours$_{\text{SSIM}}$ substantially outperforms CrossScore on the SSIM target despite a shared objective, indicating a more accurate and robust representation for structural similarity. Ours$_{\text{SSIM}}$ also performs on par with, and sometimes surpasses, the FR metric LPIPS. This demonstrates our approach can attain FR-level quality in a challenging cross-view setting without an aligned GT. %Additional experiments integrating our PR-IQA are provided in the Appendix.

Fig.~\ref{F3_Quali_Score} provides qualitative comparisons against the GT quality map (DINOv2-SIM), visually exposing the limitations of the baselines. PIQE, for instance, tends to emphasize simple edges over meaningful degradation, while PaQ-2-PiQ, PuzzleSim, and CrossScore often produce coarse, blocky maps. The patch-level design of CrossScore appears to struggle with capturing fine details and precise boundaries, resulting in coarser quality maps. By contrast, PR-IQA consistently produces maps that closely resemble the GT, recovering object silhouettes and local structural differences, thus visually validating the quantitative gains.

\begin{figure*}
    \centering
    \includegraphics[width=0.98\linewidth]{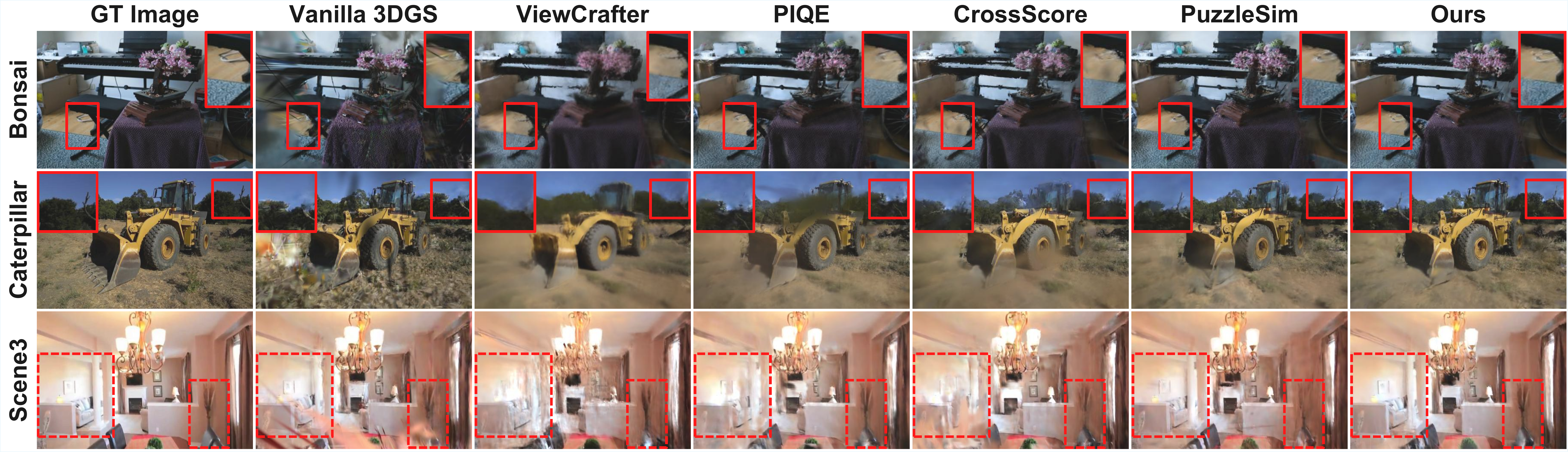}
    \caption{Qualitative comparison of rendered novel views from IQA-guided 3DGS. While baseline methods produce results with artifacts, blurring, or misaligned Gaussians, our PR-IQA-guided method (``Ours") avoids these failure modes, yielding significantly cleaner and more coherent reconstructions.}
    \label{F4_Quali_GS}
\end{figure*}

% 두 테이블 합쳐서 1 Row로
\begin{table*}
    \centering
    \begin{minipage}{0.48\textwidth}
        \centering
        \caption{Ablation on architectural components: attention block design and auxiliary inputs. PLCC $\uparrow$ and SRCC $\uparrow$ are measured against DINOv2-SIM.}
        \begin{adjustbox}{max width=\linewidth}
        \begin{tabular}{lcccc}
        \hline
        \multirow{2}{*}{Model Variants} & \multicolumn{2}{c}{Mip-NeRF 360}            & \multicolumn{2}{c}{Tanks and Temples} \\ \cline{2-5} 
                                        & \multicolumn{1}{c}{PLCC} & \multicolumn{1}{c}{SRCC} & \multicolumn{1}{c}{PLCC}   & \multicolumn{1}{c}{SRCC}   \\ \hline
        (v-i) Reversed Attention Order    & 0.540                    & 0.609                    & 0.517                      & 0.584                      \\
        (v-ii) w/o Channel Attention      & \underline {0.554}              & 0.611                    & \underline{0.571}                & 0.633                      \\
        (v-iii) w/o Reference Branch      & 0.544                    & \underline{0.613}              & 0.553                      & \underline {0.637}                \\
        (v-iv) w/o Partial Map Branch     & 0.421                    & 0.464                    & 0.452                      & 0.438                      \\
        \hline
        Full Model & \textbf{0.555}           & \textbf{0.622}           & \textbf{0.573}             & \textbf{0.650}             \\ \hline
        \end{tabular}
        \end{adjustbox}
        \label{T2_Ablation_Model}
    \end{minipage}
    \hfill
    \begin{minipage}{0.48\textwidth}
        \centering
        \caption{Quantitative comparison of IQA-guided 3DGS across IQA methods. PSNR $\uparrow$, SSIM $\uparrow$, and LPIPS $\downarrow$ are averaged over scenes. Red, orange, yellow mark 1st–3rd per column, excluding FR-based baselines$^{\dagger}$.}
        \begin{adjustbox}{max width=\linewidth}
        \begin{tabular}{llcccllllll}
        \hline
                                   &                   & \multicolumn{3}{c}{Mip-NeRF 360}                               & \multicolumn{3}{c}{Tanks and Temples}                          & \multicolumn{3}{c}{RealEstate10K}                              \\ \hline
        IQA-Guided                   & Method        & PSNR      & SSIM      & LPIPS     & \multicolumn{1}{c}{PSNR} & \multicolumn{1}{c}{SSIM} & \multicolumn{1}{c}{LPIPS} & \multicolumn{1}{c}{PSNR} & \multicolumn{1}{c}{SSIM} & \multicolumn{1}{c}{LPIPS} \\ \hline
                                   & Vanilla 3DGS              & 16.08     & 0.461     & \cellcolor[HTML]{FFDBB0}0.415     & 15.30                    & 0.509                    & \cellcolor[HTML]{FFDBB0}0.406     & 16.39                    & \cellcolor[HTML]{FFFFC7}0.625     & 0.345                     \\
        \multirow{-2}{*}{w/o IQA}        & ViewCrafter       & 16.18     & 0.474     & 0.453     & 15.77                    & 0.523                    & 0.455                     & 16.94                    & 0.620                    & \cellcolor[HTML]{FFA0A0}0.327     \\ \hline\hline
        \multirow{2}{*}{w/ FR-IQA}    & SSIM$^{\dagger}$              & 16.68     & 0.487     & 0.421     & 16.23                    & 0.556                    & 0.399                     & 17.54                    & 0.639                    & 0.325                     \\
                                   & DINOv2$^{\dagger}$            & 17.18     & 0.498     & 0.399     & 16.78                    & 0.562                    & 0.384                     & 17.83                    & 0.640                    & 0.322                     \\ \hline
                                   & PaQ-2-PiQ         & 16.30     & 0.472     & 0.425     & 15.77                    & 0.534                    & 0.421                     & 16.58                    & 0.608                    & 0.339                     \\
        \multirow{-2}{*}{w/ NR-IQA}   & PIQE              & 16.31     & 0.479     & 0.440     & 15.67                    & 0.534                    & 0.433                     & 16.54                    & 0.612                    & 0.333                     \\ \hline
                                   & CrossScore        & 16.31     & 0.476     & 0.431     & 15.86     & 0.537     & 0.427                     & \cellcolor[HTML]{FFFFC7}16.99     & 0.621                    & 0.338                     \\
                                   & PuzzleSim         & \cellcolor[HTML]{FFFFC7}16.35     & \cellcolor[HTML]{FFFFC7}0.482     & \cellcolor[HTML]{FFFFC7}0.423     & \cellcolor[HTML]{FFFFC7}15.94     & \cellcolor[HTML]{FFFFC7}0.541     & \cellcolor[HTML]{FFDBB0}0.406     & \cellcolor[HTML]{FFDBB0}17.38     & \cellcolor[HTML]{FFA0A0}0.632     & \cellcolor[HTML]{FFFFC7}0.332     \\
                                   & Ours$_{\text{SSIM}}$ & \cellcolor[HTML]{FFDBB0}16.37     & \cellcolor[HTML]{FFDBB0}0.485     & 0.427     & \cellcolor[HTML]{FFDBB0}16.14                    & \cellcolor[HTML]{FFDBB0}0.548                    & \cellcolor[HTML]{FFFFC7}0.407                     & 16.94                    & \cellcolor[HTML]{FFDBB0}0.631                    & \cellcolor[HTML]{FFDBB0}0.329                     \\
        \multirow{-4}{*}{w/ CR-IQA}   & Ours$_{\text{DINOv2}}$      & \cellcolor[HTML]{FFA0A0}16.76     & \cellcolor[HTML]{FFA0A0}0.493     & \cellcolor[HTML]{FFA0A0}0.414     & \cellcolor[HTML]{FFA0A0}16.24     & \cellcolor[HTML]{FFA0A0}0.551     & \cellcolor[HTML]{FFA0A0}0.403     & \cellcolor[HTML]{FFA0A0}17.72     & \cellcolor[HTML]{FFA0A0}0.632     & \cellcolor[HTML]{FFA0A0}0.327     \\ \hline
        \end{tabular}
        \end{adjustbox}
        \label{T3_Quanti_GS}
    \end{minipage}
    \vspace{-2mm}
\end{table*}

\subsection{Architectural Ablations}
\noindent\textbf{Setup.} We conduct a systematic ablation study to validate the core components of our model (Sect.~\ref{3.4_dense-quality}). Our analysis focuses on two key areas: (1) the design of our CBAM-like attention block, and (2) the impact of our auxiliary inputs. For the attention block, we evaluate two modified variants: (v-i) one with the spatial and channel attention modules swapped, and (v-ii) a simplified variant that removes the channel attention module, leaving a basic transformer-like structure. To assess the inputs, we test two additional variants: (v-iii) a model without the reference-image encoding branch, and (v-iv) a model without the partial quality map encoding branch. All variants are trained and evaluated under identical settings to ensure a fair comparison.

\noindent\textbf{Results.} Table~\ref{T2_Ablation_Model} reports the PLCC and SRCC for each variant on Mip-NeRF 360 and Tanks and Temples, using DINOv2-SIM as the target. The results clearly indicate that the partial quality map is the most crucial input; removing it (v-iv) causes a much larger performance drop than removing the reference image (v-iii). Furthermore, the effectiveness of our dual-gated attention block (Fig. 2(b)) is supported by the improved performance of the full model compared to the alternative variants (v-i and v-ii), with particularly notable gains in SRCC, confirming its suitability for regression of dense quality maps.

% Furthermore, the effectiveness of our dual-gated attention block (Fig.~\ref{F2_Overview}(b)) is validated by the wide margin outperformance of the full model compared to the alternative variants of the attention block (v-i and v-ii), which confirms its suitability for regression of dense quality maps.

\subsection{Application: IQA-Guided 3DGS Results}
\noindent\textbf{Setup.} We demonstrate a practical application of PR-IQA by using its predicted quality maps to guide 3DGS~\cite{kerbl20233d} training for sparse-view NVS. For each scene, we extract a GT set $\mathcal{I}_{\text{GT}}$ (100 frames for Tanks and Temples/Mip-NeRF, 25 for RealEstate10K) and select  $N_{\text{in}}$ = 5 sparse inputs ($\mathcal{I}_{\text{in}}$). For the remaining views ($\mathcal{I}_{\text{GT}} \setminus \mathcal{I}_{\text{in}}$), we use a VDM~\cite{yu2024viewcraftertamingvideodiffusion} to generate a candidate pool of $N_{v}$ = 5 images per view. 

We compare our PR-IQA against alternative IQA methods. For each IQA method, we construct a pseudo-GT set $\mathcal{I}_{\text{pseudo}}$ by selecting the highest-scoring candidate per view. The 3DGS model is trained on $\mathcal{I}_{\text{in}} \bigcup \mathcal{I}_{\text{pseudo}}$, using the method's quality maps for masking during optimization (see Sect.~\ref{4_priqa_gs}). We also evaluate two standard baselines: (i) Vanilla 3DGS~\cite{kerbl20233d}, trained only on $\mathcal{I}_{\text{in}}$, and (ii) ViewCrafter~\cite{yu2024viewcraftertamingvideodiffusion}, a diffusion baseline without IQA guidance. ViewCrafter uses the same candidate pool, randomly selects one image per view, and uses all pixels for training.

\noindent\textbf{Results.} Table~\ref{T3_Quanti_GS} presents the quantitative results. Our IQA-guided 3DGS training strategy achieves consistently strong performance, significantly outperforming both Vanilla 3DGS and ViewCrafter across all datasets. Our PR-IQA (Ours$_{\text{DINOv2}}$) achieves the highest PSNR and SSIM and the lowest LPIPS on all three datasets, indicating that its predicted quality maps effectively identify and mask inaccurate regions, thereby improving 3DGS reconstruction quality.

Fig.~\ref{F4_Quali_GS} provides a qualitative comparison for 3DGS, clearly exposing the limitations of the baseline methods. Using quality maps from other IQA methods yields views with noticeable artifacts or occlusions. The non-IQA-guided baselines also struggle: omitting diffusion-generated images entirely (Vanilla 3DGS) leads to misaligned Gaussians, while training on diffusion images without IQA guidance (ViewCrafter) produces blurry renderings. In contrast, the PR-IQA-guided 3DGS results avoid these failure modes and produce significantly cleaner, more coherent reconstructions (see red solid and dashed boxes in Fig.~\ref{F4_Quali_GS}).

\section{Conclusion}
This study introduced PR-IQA, a novel CR-IQA framework to address diffusion-generated view inconsistencies. The method computes a partial quality map in overlapping regions, then completes it into a dense map using a reference-conditioned cross-attention network. Experiments show PR-IQA outperforms existing methods, correlating highly with FR-IQA metrics without GT. Integrated into a 3DGS pipeline, its quality-aware strategy filters inconsistencies by restricting supervision to high-confidence regions, improving reconstruction fidelity. This suggests PR-IQA is a powerful tool for high-quality, sparse-view 3D reconstruction. 

\section*{Acknowledgements}
This research was supported by the MSIT(Ministry of Science and ICT), Korea, under the ITRC(Information Technology Research Center) support program(IITP-2026-RS-2020-II201789), and the Artificial Intelligence Convergence Innovation Human Resources Development(IITP-2026-RS-2023-00254592) supervised by the IITP(Institute for Information \& Communications Technology Planning \& Evaluation).

{
    \small
    \bibliographystyle{ieeenat_fullname}
    \bibliography{main}
}

\clearpage
\setcounter{page}{1}
\maketitlesupplementary

This supplementary material complements the main paper by providing comprehensive implementation details, extended experimental results, and in-depth ablation studies. Sections 1 and 2 establish the foundation for reproducibility by detailing the network architecture, loss functions, dataset generation protocols, and training configurations. We expand our experimental analysis in Section 3 to cover alternative FR-IQA targets (PSNR, LPIPS) and validate the reliability of image-level view selection. Furthermore, Sections 4 and 5 present systematic ablation studies concerning IQA design choices (e.g., reference count, fusion strategy, geometric robustness) and 3DGS parameters (e.g., guidance metric, masking threshold, soft vs. binary masking), respectively. Section 6 provides extensive qualitative visualizations for both quality map estimation and 3D reconstruction results. Finally, Section 7 discusses the limitations of the proposed method and outlines potential future directions.

\section{Method Details}

\begin{figure*}
    \centering
    \includegraphics[width=0.98\linewidth]{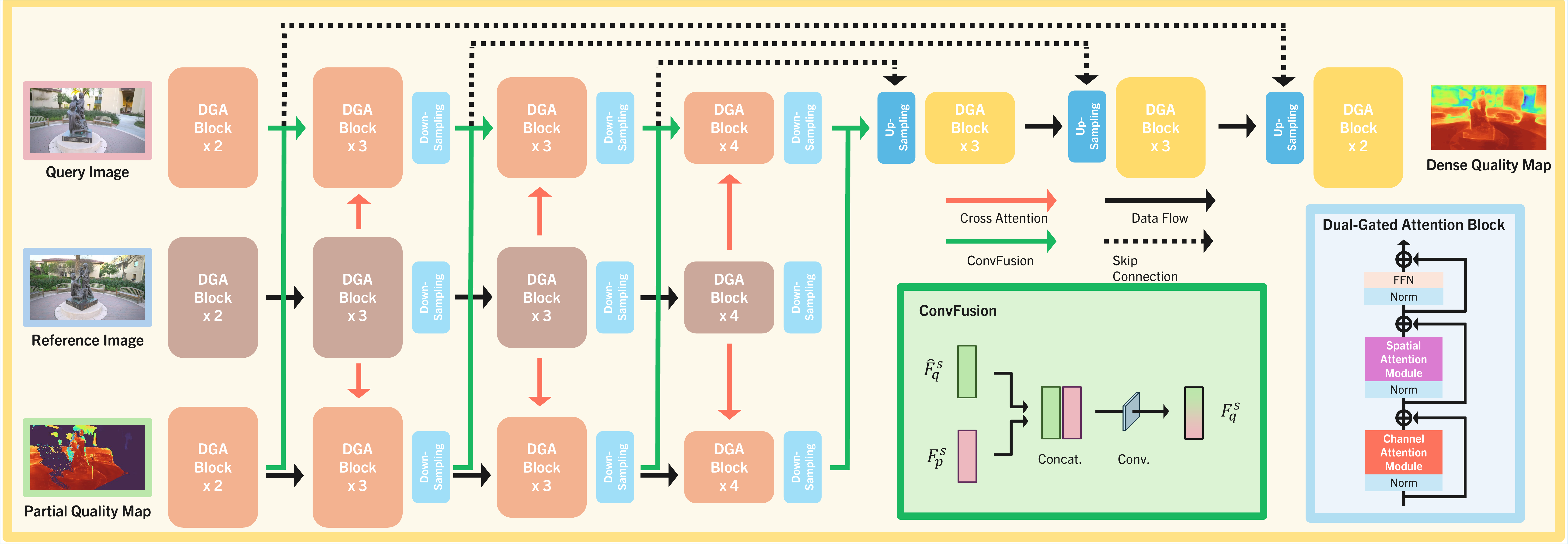}
    \vspace{-2mm}
    \caption{Detailed architecture of the proposed model. The network employs an encoder–decoder design featuring cross- and self-attention modules, query fusion, and mask-aware pixel-shuffle downsampling. Key specifications, including stage-wise block counts, attention heads, and the status of component sharing (frozen vs. trainable), are explicitly annotated.}
    \label{F1_model}
    \vspace{-2mm}
\end{figure*}

\subsection{Architecture Details}
As illustrated in Fig.~\ref{F1_model}, our architecture adopts a U-Net-like~\cite{ronneberger2015unetconvolutionalnetworksbiomedical} encoder-decoder design, leveraging DINOv2~\cite{oquab2023dinov2} as the feature backbone. The network utilizes GELU~\cite{hendrycks2023gaussianerrorlinearunits} as the activation function throughout all layers. Detailed specifications, including resolution, channel dimensions, and the number of blocks for each level, are summarized in Table~\ref{T1_architecture}.

The encoder is structured into four stages with $[2, 3, 3, 4]$ encoding blocks and $[1, 2, 4, 8]$ attention heads, respectively. The encoders for the query and reference branches share weights, whereas the encoder for the partial branch remains independent. The channel dimensions scale progressively as $[48, 96, 192, 384]$ from Level 1 to Level 4.

To effectively integrate information across branches, we employ a ConvFuse operation at each encoding stage. Specifically, the feature maps from the query and partial branches are concatenated along the channel dimension and then projected back to the original channel size via a convolutional layer. The resulting fused features serve as the input for the subsequent stage of the query branch, while the partial branch retains its original, unfused features for its own propagation.

The decoder consists of three stages containing $[3, 3, 2]$ decoding blocks and $[4, 2, 1]$ attention heads. Corresponding to the encoder levels, the decoder maintains channel widths of $192$, $96$, and $96$, respectively, following skip connection fusion and $1 \times 1$ channel reduction.

The resulting model comprises approximately 60M trainable parameters. In terms of resource consumption, it is highly efficient, requiring approximately 2 GB of GPU memory for single-image inference and 6 GB for training with a batch size of 1.

\subsection{Loss Functions}
To ensure robust performance, our training objective combines three complementary loss terms: the pixel-wise $\mathcal{L}_1$ loss for local accuracy, the Jensen-Shannon Divergence (JSD)~\cite{englesson2021generalizedjensenshannondivergenceloss} loss for global distributional alignment, and the Pearson Linear Correlation Coefficient (PLCC)~\cite{cao2022pkdgeneraldistillationframework} loss for ranking consistency.

\paragraph{Distribution Alignment (JSD Loss).} We employ the JSD loss to align the global distribution of predicted quality scores with the GT, thereby preventing mode collapse where the network predicts overly uniform values. We first flatten the quality maps $\hat{Q}$ and $Q$ into vectors $\hat{\mathbf{p}}, \mathbf{g} \in [0,1]^N$, where $N = H \times W$. Since $\hat{\mathbf{p}}$ and $\mathbf{g}$ are bounded, we first apply a logit transformation to map them into an unbounded space suitable for softmax normalization:
\begin{equation}
    \tilde{p}_i = \log\left(\frac{\hat{p}_i}{1 - \hat{p}_i}\right), \quad \tilde{g}_i = \log\left(\frac{g_i}{1 - g_i}\right).
\end{equation}
Next, we convert these logits into probability distributions $P$ and $G$ using a temperature-scaled softmax function:
\begin{equation}
    P_i = \frac{\exp(\tilde{p}_i / \tau)}{\sum_j \exp(\tilde{p}_j / \tau)}, \quad G_i = \frac{\exp(\tilde{g}_i / \tau)}{\sum_j \exp(\tilde{g}_j / \tau)}.
\end{equation}
where $\tau$ is the temperature parameter, empirically set to $0.2$. The symmetric JSD loss is then defined as the average Kullback-Leibler (KL) divergence from the mixture distribution $M = (P+G)/2$:
\begin{equation}
    \mathcal{L}_{\text{JSD}} = \frac{1}{2} \mathcal{D}_{\text{KL}}(P \parallel M) + \frac{1}{2} \mathcal{D}_{\text{KL}}(G \parallel M).
\end{equation}
This prevents mode collapse by penalizing uniform predictions: when the network predicts similar quality values everywhere, $P$ becomes nearly uniform, resulting in a large JSD loss against the typically non-uniform ground truth (GT) $G$.

\paragraph{Ranking Consistency (Pearson Loss).} To strictly enforce the relative ranking of quality, we utilize the PLCC loss. Let $\hat{\mathbf{y}}$ and $\mathbf{y}$ denote the flattened predicted and GT quality maps, respectively. We first center these vectors by subtracting their means ($\mu_{\hat{y}}, \mu_{y}$). The correlation coefficient $r$ is computed as:
\begin{equation}
    r = \frac{\sum (\hat{y}_i - \mu_{\hat{y}})(y_i - \mu_{y})}{\sqrt{\sum (\hat{y}_i - \mu_{\hat{y}})^2} \sqrt{\sum (y_i - \mu_{y})^2}}.
\end{equation}
The Pearson loss is defined as $\mathcal{L}_{\text{PLCC}} = 1 - r$. This term complements the pixel-wise $\mathcal{L}_1$ loss by focusing on linear trends and the relative ordering of salient regions, crucial for accurate quality assessment and downstream tasks, rather than solely minimizing absolute pixel errors.

\begin{table}[t]
\centering
\caption{Detailed architecture specifications of the proposed PR-IQA network. We report the spatial resolution, channel dimensions, number of attention heads, and block counts for each stage of the encoder (Enc0- Enc3) and decoder (Dec3- Dec1).}
\begin{adjustbox}{max width=\linewidth}
\begin{tabular}{lccccc}
\hline
Level & \multicolumn{1}{l}{Resolution} & \multicolumn{1}{l}{Channels} & \multicolumn{1}{l}{Heads} & \multicolumn{1}{l}{Blocks} & \multicolumn{1}{l}{Output Channels} \\ \hline
Input & 224 x 224                      & 4                            & -                         & -                          & -                                   \\
Enc0    & 224 x 224                      & 48                           & 1                         & 2                          & 48                                  \\
Enc1    & 112 x 112                      & 96                           & 2                         & 3                          & 96                                  \\
Enc2    & 56 x 56                        & 192                          & 4                         & 3                          & 192                                 \\
Enc3    & 28 x 28                        & 384                          & 8                         & 4                          & 384                                 \\
Dec3  & 56 x 56                        & 192                          & 4                         & 3                          & 192                                 \\
Dec2  & 112 x 112                      & 96                           & 2                         & 3                          & 96                                  \\
Dec1  & 224 x 224                      & 96                           & 1                         & 2                          & 96                                  \\
Output& 224 x 224                      & 1                            & -                         & -                          & 1                                   \\ \hline
\end{tabular}
\end{adjustbox}
\label{T1_architecture}
\end{table}

\section{Experimental Details}

\subsection{Training Data Generation}
\paragraph{Frame Sampling.}
We utilize the Map-free Visual Relocalization (MFR) dataset~\cite{arnold2022map} as our primary source. For each scene, we uniformly sample 200 frames along the camera trajectory, explicitly including the start and end frames. This uniform sampling strategy serves two purposes: it reduces the computational overhead for the Video Diffusion Model (VDM)~\cite{yu2024viewcraftertamingvideodiffusion} and prevents redundancy by mitigating negligible pose changes between adjacent frames.

\begin{table}[t]                                
    \centering
    \small
    \caption{List of evaluation scenes. We enumerate the specific scenes and sequence IDs selected from the Mip-NeRF 360, Tanks and Temples, and RealEstate10K datasets used for our experimental benchmarks.}
    \begin{adjustbox}{max width=\linewidth}
    \begin{tabular}{c|cccccc}
        \hline
        Dataset & \multicolumn{6}{c}{Scene} \\
        \hline
        Mip-NeRF 360 
            & Bonsai & Counter & Garden & Kitchen & Room & Treehill \\
        \hline
        Tanks and Temples
            & Barn & Caterpillar & Family & Horse & Ignatius & Truck \\
        \hline
        \multirow{2}{*}{RealEstate10K}
            & \multicolumn{2}{c}{87f03b8928fc286e}
            & \multicolumn{2}{c}{d932fa3862974507}
            & \multicolumn{2}{c}{9ea61697c238be3d} \\
        & \multicolumn{2}{c}{7bab7b21dbaf38ab}
            & \multicolumn{2}{c}{2e7ffcba51990c93}
            & \multicolumn{2}{c}{f48829b917629fe0} \\
        \hline
    \end{tabular}
    \end{adjustbox}
    \label{T2_scene_detail}
\end{table}

\paragraph{View Synthesis and Distortion.}
Following the ViewCrafter protocol~\cite{yu2024viewcraftertamingvideodiffusion}, we organize the sampled frames into sliding windows of size 25. Within each window, two anchor images are used to synthesize novel views. It is a known characteristic of VDMs that generation fidelity degrades as the target viewpoint deviates further from the conditioning camera poses. We explicitly leverage this property to induce a diverse spectrum of realistic artifacts and geometric distortions in the generated images. This strategy enriches our training distribution with challenging samples, thereby enhancing the model's robustness to reconstruction errors.

\begin{table*}[t]
\centering
% \small
\caption{Quantitative comparisons of predicted quality maps against GT quality maps (PLCC↑, SRCC↑), targeting PSNR and LPIPS. Red, orange, and yellow cells denote the 1st, 2nd, and 3rd best methods per column (excluding FR settings †), while gray cells indicate identity cases where the IQA prediction matches the GT quality map. }
\begin{adjustbox}{max width=\linewidth}
\begin{tabular}{llcccccccccccc}
\hline
                           &                         & \multicolumn{4}{c}{Mip-NeRF 360}                                           & \multicolumn{4}{c}{Tanks and Temples}                                      & \multicolumn{4}{c}{RealEstate10K}                                          \\ \cline{3-14}
IQA Type                   & IQA Method                & \multicolumn{2}{c}{PSNR}      & \multicolumn{2}{c}{LPIPS}    & \multicolumn{2}{c}{PSNR}      & \multicolumn{2}{c}{LPIPS}    & \multicolumn{2}{c}{PSNR}      & \multicolumn{2}{c}{LPIPS}    \\ \cline{3-14}
                           &                           & PLCC          & SRCC          & PLCC          & SRCC          & PLCC          & SRCC          & PLCC          & SRCC          & PLCC          & SRCC          & PLCC          & SRCC          \\ \hline
                           & PSNR$^\dagger$                      & \cellcolor[HTML]{C0C0C0}1.000         & \cellcolor[HTML]{C0C0C0}1.000         & 0.434         & 0.384         & 1.000         & 1.000         & 0.478         & 0.459         & \cellcolor[HTML]{C0C0C0}1.000         & \cellcolor[HTML]{C0C0C0}1.000         & 0.370         & 0.347         \\
                           & SSIM$^\dagger$                      & 0.517         & 0.487         & 0.565         & 0.554         & 0.486         & 0.487         & 0.598         & 0.595         & 0.392         & 0.386         & 0.452         & 0.460         \\
                           & LPIPS$^\dagger$                     & 0.434         & 0.384         & \cellcolor[HTML]{C0C0C0}1.000         & \cellcolor[HTML]{C0C0C0}1.000         & 0.478         & 0.459         & \cellcolor[HTML]{C0C0C0}1.000         & \cellcolor[HTML]{C0C0C0}1.000         & 0.370         & 0.347         & \cellcolor[HTML]{C0C0C0}1.000         & \cellcolor[HTML]{C0C0C0}1.000         \\
\multirow{-4}{*}{FR-IQA} & DINOv2$^\dagger$                    & 0.407         & 0.338         & 0.557         & 0.472         & 0.396         & 0.361         & 0.582         & 0.581         & 0.248         & 0.241         & 0.489         & 0.516         \\ \hline \hline
                           & PAL4VST                   & 0.016         & 0.016         & 0.024         & 0.021         & 0.004         & 0.004         & 0.004         & 0.004         & 0.013         & 0.012         & 0.078         & 0.074         \\
                           & PaQ-2-PiQ                 & -0.179        & -0.181        & -0.047        & -0.047        & -0.136        & -0.095        & 0.007         & 0.053         & -0.126        & -0.134        & 0.029         & 0.030         \\
\multirow{-3}{*}{NR-IQA}   & PIQE                      & -0.110        & -0.114        & 0.031         & 0.035         & 0.223         & 0.242         & 0.194         & 0.208         & \cellcolor{MyOrange}0.227         & \cellcolor{MyOrange}0.235         & 0.047         & 0.062         \\ \hline
                           & MEt3R* & 0.056         & 0.055         & 0.057         & 0.042         & 0.106         & 0.120         & 0.181         & 0.196         & 0.125         & 0.117         & \cellcolor{MyOrange}0.363         & \cellcolor{MyOrange}0.352         \\
                           & CrossScore                & 0.082         & 0.081         & 0.224         & \cellcolor{MyOrange}0.238         & 0.206         & 0.182         & 0.312         & 0.304         & 0.195         & 0.149         & 0.169         & 0.161         \\
                           & PuzzleSim                 & \cellcolor{MyYellow}0.179         & 0.172         & \cellcolor{MyRed}0.286         & \cellcolor{MyRed}0.264         & \cellcolor{MyYellow}0.250         & \cellcolor{MyOrange}0.259         & \cellcolor{MyRed}0.456         & \cellcolor{MyRed}0.433         & \cellcolor{MyYellow}0.208         & 0.200         & \cellcolor{MyRed}0.458         & \cellcolor{MyRed}0.447         \\
                           & Ours$_{\text{partial}}$* & 0.161         & \cellcolor{MyYellow}0.184         & 0.189         & 0.173         & 0.131         & 0.134         & 0.225         & 0.256         & 0.070         & 0.150         & 0.208         & 0.298         \\
                           & Ours$_{\text{DINOv2}}$      & \cellcolor{MyOrange}0.259         & \cellcolor{MyOrange}0.227         & \cellcolor{MyOrange}0.280         & 0.229         & \cellcolor{MyOrange}0.273         & \cellcolor{MyYellow}0.258         & \cellcolor{MyOrange}0.401         & \cellcolor{MyOrange}0.384         & 0.206         & \cellcolor{MyYellow}0.215         & \cellcolor{MyYellow}0.304         & \cellcolor{MyYellow}0.333         \\
\multirow{-6}{*}{CR-IQA}   & Ours$_{\text{SSIM}}$       & \cellcolor{MyRed}0.338         & \cellcolor{MyRed}0.345         & \cellcolor{MyYellow}0.235         & \cellcolor{MyYellow}0.229         & \cellcolor{MyRed}0.340         & \cellcolor{MyRed}0.334         & \cellcolor{MyYellow}0.340         & \cellcolor{MyYellow}0.334         & \cellcolor{MyRed}0.284         & \cellcolor{MyRed}0.244         & 0.171         & 0.175         \\ \hline
\end{tabular}
\end{adjustbox}
\begin{tablenotes}
\footnotesize
\item $\dagger$ Metrics require a same-pose GT image.  * Metrics are computed only over the valid overlapping region.
\end{tablenotes}
\label{T3_PSNR_LPIPS}
\end{table*}

\paragraph{Reference Selection and Annotation.}
To ensure sufficient baseline separation and avoid trivial correlations from high-overlap pairs, we systematically select reference frames relative to the query. For a given query frame $I_q$, we identify four reference candidates $\{I_r\}$ at relative indices of $\pm10$ and $\pm20$ within the sampled sequence. For each resulting query-reference pair $(I_q, I_r)$, we generate pseudo-ground-truth supervision by applying the procedure described in the Partial Map Generation section (Sect. 3.3 of the main manuscript). This involves estimating global point clouds via dense stereo matching, performing z-buffered reprojection to align views, and finally computing the partial quality map $\hat{Q}$.

\paragraph{Data Structure.}
Consequently, training samples are formed as tuples $(I_q, I_r, \hat{Q}, Q^\ast)$, where $Q^\ast$ represents the GT quality map. This structure enables a systematic evaluation of robustness in the CR-IQA setting.

\subsection{Evaluation Data Generation}
\paragraph{Dataset Selection.}
We conduct our evaluation across three standard benchmarks: Mip-NeRF 360, Tanks and Temples, and RealEstate10K. The specific scenes selected for these experiments are listed in Table~\ref{T2_scene_detail} (with RealEstate10K sequences indexed as Scenes 1–6). To ensure consistency, we employ the identical set of scenes for both the standalone IQA performance assessment and the downstream IQA-guided 3DGS experiments.

\paragraph{Query Image Synthesis.}
To generate the synthesized query images used for evaluation, we adopt a standardized pipeline. We utilize the sequence endpoints (i.e., the first and last frames) as the reference views. For the intermediate target frames, we employ DUSt3R~\cite{wang2024dust3rgeometric3dvision} to estimate dense point clouds by matching the endpoints with the current frame. These point clouds are subsequently rendered into the target viewpoint and processed by the VDM to refine the details, producing the final query images.

\subsection{Model Training}
All input images are resized to a resolution of $294 \times 518$. The model is trained using the AdamW optimizer~\cite{loshchilov2017sgdrstochasticgradientdescent} with $\beta_1 = 0.9$ and $\beta_2 = 0.999$, starting with an initial learning rate of $1 \times 10^{-4}$. We employ a Cosine Annealing with Warm Restarts schedule~\cite{loshchilov2017sgdrstochasticgradientdescent}, where the learning rate decays to $1 \times 10^{-6}$ with a restart period of 135,000 iterations. The entire training process spans 270,000 iterations (approximately 20 hours) on four NVIDIA RTX 3090 GPUs, utilizing a total batch size of 12 (3 frames per GPU).

\begin{table*}[t]
\centering
\caption{Image selection evaluation. We report the correlation (PLCC, SRCC) between per-image quality scores and ground-truth quality scalars derived from DINOv2 feature similarity and SSIM across three datasets. Ours$_{\text{DINOv2}}$  demonstrates strong alignment with feature-based quality, achieving the highest performance on Tanks and Temples and RealEstate10K, and competitive results on Mip-NeRF 360.}
\begin{adjustbox}{max width=\linewidth}
\begin{tabular}{llccccccccccccc}
\hline
                         &              & \multicolumn{4}{c}{Mip-NeRF 360}                                                                                                                                                                                                  & \multicolumn{4}{c}{Tanks and Temples}                                                                                                                                                                                             & \multicolumn{4}{c}{RealEstate10K}                                                                                                                                                                                                 \\ \hline
IQA Type                 & IQA Method   & \begin{tabular}[c]{@{}c@{}}PLCC\\ (DINOv2)\end{tabular} & \begin{tabular}[c]{@{}c@{}}SRCC\\ (DINOv2)\end{tabular} & \begin{tabular}[c]{@{}c@{}}PLCC\\ (SSIM)\end{tabular} & \begin{tabular}[c]{@{}c@{}}SRCC\\ (SSIM)\end{tabular} & \begin{tabular}[c]{@{}c@{}}PLCC\\ (DINOv2)\end{tabular} & \begin{tabular}[c]{@{}c@{}}SRCC\\ (DINOv2)\end{tabular} & \begin{tabular}[c]{@{}c@{}}PLCC\\ (SSIM)\end{tabular} & \begin{tabular}[c]{@{}c@{}}SRCC\\ (SSIM)\end{tabular} & \begin{tabular}[c]{@{}c@{}}PLCC\\ (DINOv2)\end{tabular} & \begin{tabular}[c]{@{}c@{}}SRCC\\ (DINOv2)\end{tabular} & \begin{tabular}[c]{@{}c@{}}PLCC\\ (SSIM)\end{tabular} & \begin{tabular}[c]{@{}c@{}}SRCC\\ (SSIM)\end{tabular} \\ \hline
                         & PaQ-2-PiQ      & 0.002                                                   & 0.012                                                   & -0.014                                                & -0.009                                                & 0.113                                                   & 0.112                                                   & -0.166                                                & -0.179                                                & 0.022                                                   & 0.012                                                   & 0.032                                                 & 0.042                                                 \\
\multirow{-2}{*}{NR-IQA} & PIQE         & 0.047                                                   & 0.044                                                   & 0.348                                                 & 0.347                                                 & 0.075                                                   & 0.075                                                   & 0.329                                                 & 0.385                                                 & -0.126                                                  & -0.128                                                  & -0.133                                                & -0.118                                                \\ \hline
                         & CrossScore & -0.090                                                  & -0.104                                                  & 0.366                                                 & 0.366                                                 & 0.188                                                   & 0.188                                                   & -0.097                                                & -0.126                                                & -0.095                                                  & -0.074                                                  & -0.035                                                & -0.026                                                \\
                         & PuzzleSim  & \cellcolor{MyRed}0.607                           & \cellcolor{MyOrange}0.518                           & \cellcolor{MyYellow}0.516                         & \cellcolor{MyYellow}0.494                         & \cellcolor{MyOrange}0.616                           & \cellcolor{MyOrange}0.612                           & \cellcolor{MyOrange}0.539                         & \cellcolor{MyYellow}0.399                         & \cellcolor{MyOrange}0.772                           & \cellcolor{MyOrange}0.727                           & \cellcolor{MyRed}0.827                         & \cellcolor{MyOrange}0.747                         \\
                         & Ours$_{\text{SSIM}}$    & \cellcolor{MyYellow}0.186                           & \cellcolor{MyYellow}0.164                           & \cellcolor{MyRed}0.629                         & \cellcolor{MyRed}0.620                         & \cellcolor{MyYellow}0.278                           & \cellcolor{MyYellow}0.287                           & \cellcolor{MyYellow}0.457                         & \cellcolor{MyOrange}0.511                         & \cellcolor{MyYellow}0.595                           & \cellcolor{MyYellow}0.600                           & \cellcolor{MyYellow}0.666                         & \cellcolor{MyYellow}0.684                         \\
\multirow{-4}{*}{CR-IQA} & Ours$_{\text{DINOv2}}$    & \cellcolor{MyOrange}0.597                           & \cellcolor{MyRed}0.547                           & \cellcolor{MyOrange}0.571                         & \cellcolor{MyOrange}0.541                         & \cellcolor{MyRed}0.627                           & \cellcolor{MyRed}0.619                           & \cellcolor{MyRed}0.590                         & \cellcolor{MyRed}0.557                         & \cellcolor{MyRed}0.790                           & \cellcolor{MyRed}0.802                           & \cellcolor{MyOrange}0.746                         & \cellcolor{MyRed}0.783                         \\ \hline
\end{tabular}
\end{adjustbox}
\label{T4_image_selection}
\end{table*}

\subsection{Baseline Details}
We compare our method against a comprehensive set of baselines across three categories: Full-Reference (FR), No-Reference (NR), and Cross-Reference (CR) IQA.

\begin{itemize}
    \item \textbf{FR-IQA:} We utilize PSNR and SSIM~\cite{wang2004image} as representative metrics for measuring pixel-wise reconstruction error and structural similarity, respectively. Additionally, LPIPS~\cite{Zhang_2018_CVPR} is employed to assess perceptual similarity based on deep feature distances extracted from pre-trained networks.
    
    \item \textbf{NR-IQA:} PAL4VST~\cite{zhang2023perceptual} is a segmentation-based model trained on pixel-level artifact masks. PaQ-2-PiQ~\cite{ying2020patches} uses a ResNet-based~\cite{he2015deepresiduallearningimage} architecture to jointly learn local (patch-level) and global (image-level) quality. PIQE~\cite{venkatanath2015blind} is a training-free method that quantifies distortions, such as blur and noise, by analyzing the statistical properties of spatially active blocks.
    
    \item \textbf{CR-IQA:} MEt3R~\cite{asim2025met3r} evaluates multi-view consistency by using dense stereo to project DINO~\cite{caron2021emerging} and FeatUp~\cite{fu2024featupmodelagnosticframeworkfeatures} features into a shared 3D space, followed by cosine similarity computation. CrossScore~\cite{wang2024crossscore} utilizes a DINOv2~\cite{oquab2023dinov2} encoder with a cross-attention module to compare the query against multiple references, predicting a patch-level map approximating SSIM. PuzzleSim~\cite{hermann2025puzzle} operates in the feature space of a pre-trained network, producing a similarity map based on patch statistics learned from scene training views.
\end{itemize}

For all learning-based baselines, we use the publicly available pre-trained models without additional fine-tuning.

\section{More Experimental Results}

\subsection{Evaluation on Alternative FR-IQA Targets}
Although our Partial-Reference (PR-IQA) framework is trained to optimize DINOv2-SIM and SSIM maps, we extend our evaluation to alternative FR-IQA targets, specifically PSNR and LPIPS, to assess the generalization capability of our predicted quality maps. Table~\ref{T3_PSNR_LPIPS} summarizes the Pearson Linear Correlation Coefficient (PLCC) and Spearman Rank Correlation Coefficient (SRCC) between our predicted maps $Q$ and the GT quality maps $Q\ast$ derived from these unseen metrics.

\paragraph{Evaluation on PSNR.}
As shown in Table~\ref{T3_PSNR_LPIPS}, our method demonstrates robust generalization to the PSNR target, which measures pixel-level fidelity. Specifically for the PSNR target, $\text{Ours}_{\text{SSIM}}$ achieves state-of-the-art performance, ranking first across all datasets. $\text{Ours}_{\text{DINOv2}}$ also shows competitive correlations, generally outperforming other baselines. In stark contrast, NR-IQA baselines (PAL4VST, PaQ-2-PiQ, PIQE) exhibit extremely low or even negative correlations. This suggests that traditional natural-image quality predictors fail to capture the specific rendering artifacts inherent in novel view synthesis. While some CR-IQA methods like PuzzleSim show moderate success, our method proves significantly more effective at approximating the pixel-wise accuracy required for PSNR prediction.

\paragraph{Evaluation on LPIPS.}
For the LPIPS target, the CR-IQA baseline PuzzleSim generally ranks first. This performance is likely attributable to architectural bias: PuzzleSim relies on VGG features, which structurally align with the VGG backbone used in LPIPS. Despite this advantage, $\text{Ours}_{\text{DINOv2}}$ achieves highly competitive results, consistently ranking second on Mip-NeRF 360 and Tanks and Temples. This indicates that our method effectively captures perceptual quality variations even without relying on the same feature backbone as the target metric. Other CR-IQA methods (MEt3R, CrossScore) show lower correlations, and NR-IQA methods again fail to provide meaningful estimates.

Our approach demonstrates superior generalization compared to existing methods. $\text{Ours}_{\text{SSIM}}$ and $\text{Ours}_{\text{DINOv2}}$ effectively generalize PSNR and LPIPS targets respectively, significantly outperforming NR-IQA. Furthermore, compared to CR-IQA baselines, our strategy of learning quality completion from partial references proves to be a more robust solution for estimating diverse quality metrics.

\begin{figure*}[t]
    \centering
    \includegraphics[width=0.98\linewidth]{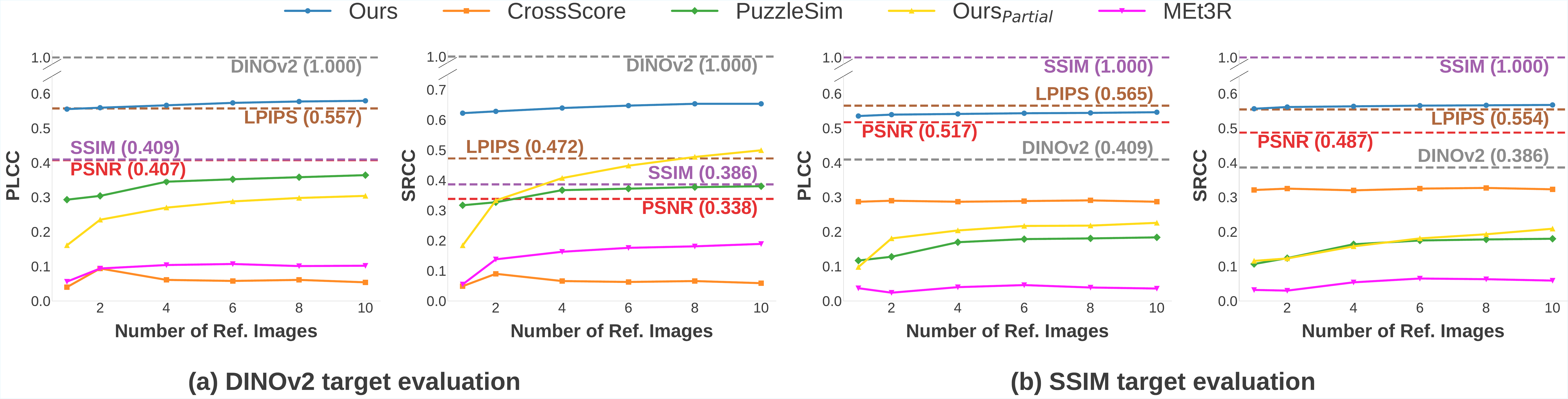}
    \vspace{-2mm}
    \caption{Impact of the number of reference views on IQA performance. We plot the PLCC and SRCC against the number of reference images used for evaluation. FR-IQA baselines are indicated by constant horizontal lines. The results are shown for (a) DINOv2 and (b) SSIM targets. In both scenarios, our model achieves the highest correlation among learned metrics (CrossScore, PuzzleSim) and demonstrates robustness even with a single reference image.}
    \label{F2_ref_num}
    \vspace{-2mm}
\end{figure*}

\begin{figure}[t]
    \centering
    \includegraphics[width=0.98\linewidth]{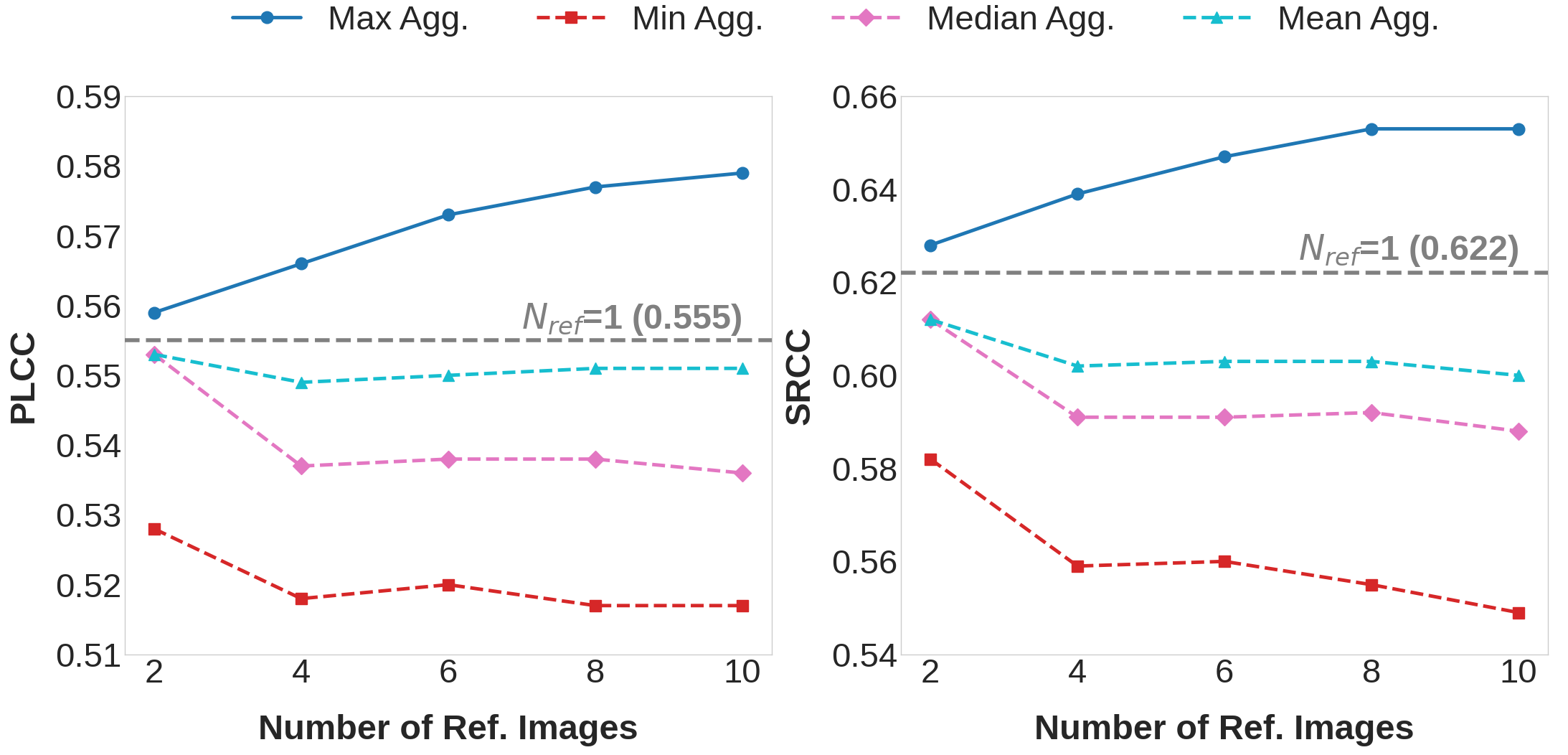}
    \vspace{-2mm}
    \caption{Impact of quality map fusion strategies on DINOv2 target evaluation. We evaluate the performance of four aggregation strategies (Max, Min, Median, and Mean) as a function of the number of reference images ($N_{\text{ref}} \in [2, 10]$).}
    \label{F3_fusion}
    \vspace{-2mm}
\end{figure}

\subsection{Evaluation on Image Selection for 3DGS}

We employ an image-level quality score to select the optimal pseudo-ground-truth candidate from the diffusion-generated pool (in Sect.~4 of the main manuscript). In this section, we quantitatively evaluate the reliability of various IQA metrics for this selection task.

To validate whether image-level scores effectively represent semantic quality, we analyze the correlation between the predicted scores and the GT DINOv2 feature similarity maps. Specifically, for each generated image, we compute the pixel-wise cosine similarity between its DINOv2 features and those of the corresponding real image at the same pose. This dense similarity map is then spatially averaged to derive a single scalar GT score. We verify the alignment by measuring the PLCC and SRCC correlations between this scalar and the scores predicted by different IQA methods (CR-IQA and NR-IQA) across all test frames.

Table~\ref{T4_image_selection} presents the correlation results on three benchmark datasets. $\text{Ours}_{\text{DINOv2}}$ demonstrates robust and consistent performance across all datasets. It generally achieves the highest correlations on Tanks and Temples and RealEstate10K, significantly outperforming baselines. On Mip-NeRF 360, it remains highly competitive, showing results comparable to PuzzleSim. While PuzzleSim also exhibits strong correlations thanks to its VGG-based feature representation, our method proves to be more effective in scenarios requiring precise semantic alignment, such as RealEstate10K.
 
In stark contrast, NR-IQA methods (PaQ-2-PiQ, PIQE) exhibit weak or near-zero correlations across all datasets. This indicates that no-reference metrics, which focus on low-level perceptual artifacts, fail to capture the reference-relative semantic quality required for 3DGS. Similarly, CrossScore displays inconsistent behavior, yielding negative correlations on Mip-NeRF 360, suggesting that its matching-based mechanism does not reliably align with dense feature similarity.

\begin{table*}[t]
\centering
\caption{Cross-generator generalization results on two unseen generators, GEN3C and SEVA, evaluated without any retraining. We report PLCC and SRCC on Mip-NeRF 360 and Tanks and Temples using DINOv2- and SSIM-based target quality scores.}
\setlength{\tabcolsep}{2.5pt}
\begin{adjustbox}{max width=0.95\linewidth}
\begin{threeparttable}
\begin{tabular}{lcccccccccccccccc}
\toprule
\multicolumn{1}{c}{} &
\multicolumn{4}{c}{Mip-NeRF 360 (GEN3C)} &
\multicolumn{4}{c}{Tanks and Temples (GEN3C)} &
\multicolumn{4}{c}{Mip-NeRF (SEVA)} &
\multicolumn{4}{c}{Tanks and Temples (SEVA)} \\
\cmidrule(lr){2-5}\cmidrule(lr){6-9}\cmidrule(lr){10-13}\cmidrule(lr){14-17}

\multicolumn{1}{c}{} &
\multicolumn{2}{c}{DINOv2} & \multicolumn{2}{c}{SSIM} &
\multicolumn{2}{c}{DINOv2} & \multicolumn{2}{c}{SSIM} &
\multicolumn{2}{c}{DINOv2} & \multicolumn{2}{c}{SSIM} &
\multicolumn{2}{c}{DINOv2} & \multicolumn{2}{c}{SSIM} \\
\cmidrule(lr){2-3}\cmidrule(lr){4-5}
\cmidrule(lr){6-7}\cmidrule(lr){8-9}
\cmidrule(lr){10-11}\cmidrule(lr){12-13}
\cmidrule(lr){14-15}\cmidrule(lr){16-17}

\multicolumn{1}{c}{\multirow{-3}{*}{IQA Method}} &
\multicolumn{1}{c}{PLCC} & \multicolumn{1}{c}{SRCC} &
\multicolumn{1}{c}{PLCC} & \multicolumn{1}{c}{SRCC} &
\multicolumn{1}{c}{PLCC} & \multicolumn{1}{c}{SRCC} &
\multicolumn{1}{c}{PLCC} & \multicolumn{1}{c}{SRCC} &
\multicolumn{1}{c}{PLCC} & \multicolumn{1}{c}{SRCC} &
\multicolumn{1}{c}{PLCC} & \multicolumn{1}{c}{SRCC} &
\multicolumn{1}{c}{PLCC} & \multicolumn{1}{c}{SRCC} &
\multicolumn{1}{c}{PLCC} & \multicolumn{1}{c}{SRCC} \\
\midrule

MEt3R* & 0.251 & \cellcolor{MyYellow}0.279 & 0.070 & 0.061 & 0.254 & 0.231 & 0.121 & 0.121 & 0.168 & 0.227 & 0.086 & 0.098 & 0.142 & 0.144 & 0.120 & 0.141 \\
CrossScore & 0.076 & 0.092 & \cellcolor{MyYellow}0.229 & \cellcolor{MyYellow}0.220 & \cellcolor{MyYellow}0.345 & 0.365 & \cellcolor{MyOrange}0.530 & \cellcolor{MyOrange}0.520 & -0.005 & 0.026 & \cellcolor{MyYellow}0.187 & \cellcolor{MyYellow}0.185 & 0.204 & 0.276 & \cellcolor{MyOrange}0.395 & \cellcolor{MyOrange}0.381 \\
PuzzleSim & \cellcolor{MyYellow}0.258 & 0.271 & 0.153 & 0.153 & \cellcolor{MyOrange}0.422 & \cellcolor{MyOrange}0.435 & \cellcolor{MyYellow}0.420 & \cellcolor{MyYellow}0.413 & \cellcolor{MyOrange}0.312 & \cellcolor{MyYellow}0.338 & 0.160 & 0.164 & \cellcolor{MyOrange}0.331 & \cellcolor{MyYellow}0.367 & \cellcolor{MyYellow}0.327 & \cellcolor{MyYellow}0.320 \\
Ours$_{\text{partial}}$* & \cellcolor{MyOrange}0.308 & \cellcolor{MyRed}0.409 & 0.174 & 0.178 & 0.344 & \cellcolor{MyYellow}0.433 & 0.067 & 0.084 & \cellcolor{MyYellow}0.258 & \cellcolor{MyOrange}0.409 & 0.119 & 0.164 & \cellcolor{MyYellow}0.318 & \cellcolor{MyOrange}0.504 & 0.087 & 0.113 \\
Ours$_{\text{DINOv2}}$ & \cellcolor{MyRed}0.368 & \cellcolor{MyOrange}0.401 & \cellcolor{MyOrange}0.303 & \cellcolor{MyOrange}0.287 & \cellcolor{MyRed}0.548 & \cellcolor{MyRed}0.596 & 0.392 & 0.403 & \cellcolor{MyRed}0.358 & \cellcolor{MyRed}0.472 & \cellcolor{MyOrange}0.306 & \cellcolor{MyOrange}0.313 & \cellcolor{MyRed}0.418 & \cellcolor{MyRed}0.543 & 0.299 & 0.276 \\
Ours$_{\text{SSIM}}$ & 0.113 & 0.136 & \cellcolor{MyRed}0.340 & \cellcolor{MyRed}0.341 & 0.328 & 0.340 & \cellcolor{MyRed}0.558 & \cellcolor{MyRed}0.553 & 0.085 & 0.143 & \cellcolor{MyRed}0.431 & \cellcolor{MyRed}0.420 & 0.211 & 0.294 & \cellcolor{MyRed}0.547 & \cellcolor{MyRed}0.521 \\
\bottomrule
\end{tabular}

\end{threeparttable}
\end{adjustbox}
\label{tab:unseen_generators}
\end{table*}

\subsection{Generalization to Unseen Generators}
\label{sec:unseen_generators}

To evaluate cross-generator generalization, we applied PR-IQA directly to images synthesized by unseen generators (\textbf{GEN3C~\cite{ren2025gen3c}} and \textbf{SEVA~\cite{zhou2025stablevirtualcameragenerative}}) without any retraining. As shown in Table~\ref{tab:unseen_generators}, the evaluation demonstrates that our model maintains a stable correlation across various datasets and target metrics, indicating that the learned quality cues are not tied to the rendering characteristics of a specific generator. Unlike prior methods that exhibit significant performance fluctuations depending on the generator or evaluation metric, PR-IQA consistently yields competitive and superior results. This suggests that our partial-reference formulation effectively captures transferable perceptual correspondences rather than overfitting to generator-specific artifacts, thereby demonstrating robust generalization capabilities on previously unseen generative models.

\section{More Ablation Studies on IQA}

\subsection{Impact of the Number of Reference Images}
We conducted an ablation study to analyze the sensitivity of our PR-IQA framework to the number of available reference images $N_{\text{ref}}$. In this experiment, we varied $N_{\text{ref}}$ from 1 to 10 by selecting reference views at regular intervals from the corresponding image sequence.

Fig.~\ref{F2_ref_num} illustrates the evolution of PLCC and SRCC scores for both DINOv2 and SSIM targets as the number of reference views increases. In contrast to CrossScore, where performance saturates, all other methods exhibit a steady gain in performance with additional reference views. Notably, our models ($\text{Ours}_{\text{DINOv2}}$ and $\text{Ours}_{\text{SSIM}}$) demonstrate high robustness even with a single reference view and continue to improve monotonically.

A significant finding is that our method achieves parity with, or even surpasses, established FR metrics without requiring GT supervision. Specifically, as shown in the DINOv2 target evaluation, our method begins to outperform the LPIPS baseline (orange dotted line) once $N_{\text{ref}} \ge 4$. This confirms that with sufficient cross-view context, our framework can predict quality maps with FR-level accuracy.

The $\hat{Q}$ variant (yellow solid line), which relies solely on geometrically overlapping regions, shows a steep performance increase as $N_{\text{ref}}$ grows. This trend validates our design rationale: increasing the number of reference views expands the geometric coverage of the partial quality map $\hat{Q}$, thereby providing a richer guidance signal for the subsequent quality completion network.

\subsection{Quality Fusion Strategy}
We investigate the optimal strategy for aggregating quality predictions when multiple reference images are available. As illustrated in Fig.~\ref{F3_fusion}, we evaluate four pixel-wise fusion operators, Max, Min, Median, and Mean, across varying reference counts ($N_{\text{ref}} = 1$ to $10$) to determine the most effective aggregation method.

Given $K$ reference images yielding predicted quality maps $\{Q_i\}_{i=1}^K$ for a query image $I_q$, the fused map values at pixel $p$ are computed as follows:

\begin{equation}
    \begin{aligned}
        Q_{\text{max}}(p) &= \max_{i} \{ Q_i(p) \}, \\
        Q_{\text{min}}(p) &= \min_{i} \{ Q_i(p) \}, \\
        Q_{\text{mean}}(p) &= \frac{1}{K} \sum_{i=1}^{K} Q_i(p), \\
        Q_{\text{median}}(p) &= \text{median}_{i} \{ Q_i(p) \}.
    \end{aligned}
\end{equation}

\begin{table}[t]
\centering
\small
\caption{Ablation study on the contribution of loss components. We compare the full model with variants trained without the JSD loss (w/o $\mathcal{L}_\text{JSD}$) or without the PLCC loss (w/o $\mathcal{L}_\text{PLCC}$). All metrics are evaluated on the Mip-NeRF 360 and Tanks and Temples datasets using PLCC and SRCC for the target of DINOv2. Bold indicates the best performance.}
\begin{tabular}{lcccc}
\hline
\multirow{2}{*}{Loss variants} & \multicolumn{2}{c}{Mip-NeRF 360} & \multicolumn{2}{c}{Tanks and Temples} \\ \cline{2-5} 
                               & PLCC            & SRCC           & PLCC              & SRCC              \\ \hline
w/o $\mathcal{L}_{\text{JSD}}$                   & -0.181          & -0.202         & -0.242            & -0.274            \\
w/o $\mathcal{L}_{\text{PLCC}}$                 & -0.119          & -0.134         & -0.147            & -0.150            \\
Full Model                     & \textbf{0.555}           & \textbf{0.622}          & \textbf{0.573}             & \textbf{0.649}             \\ \hline
\end{tabular}
\label{T5_loss}
\end{table}

The quantitative results demonstrate that the Max fusion strategy consistently outperforms all other aggregation methods. As shown in Fig.~\ref{F3_fusion}, the performance of Max fusion improves monotonically as the number of reference images increases, reaching peak correlations at $N_{\text{ref}} = 10$. This represents a substantial gain over the single-reference baseline.

In contrast, Min fusion exhibits the poorest performance, showing a degrading trend where accuracy drops significantly as more references are added. The Mean and Median strategies remain relatively stagnant and fail to consistently surpass the single-reference baseline.

The widening gap between Max fusion and other methods suggests that an optimistic aggregation strategy is crucial for robust cross-reference evaluation. By selecting the maximum quality score per pixel, the framework effectively isolates the best matching evidence from the available views. This approach allows the model to filter out low scores caused by occlusions, view-dependent artifacts, or poor geometric correspondences in specific reference frames, ensuring that the final quality map reflects the most reliable visual information.

\begin{table*}[t]
\centering
\small
\caption{Geometric robustness analysis under point cloud filtering and camera pose perturbations. We evaluate the sensitivity of our PR-IQA framework to geometric input quality on the Mip-NeRF 360 and Tanks and Temples datasets. We analyze the impact of varying VGGT depth confidence filtering thresholds (No filtering, 20\%, 50\%) and introduce synthetic Gaussian noise to camera parameters (5\% and 10\% levels). Red, orange, and yellow cells denote the 1st, 2nd, and 3rd best results, respectively. The results demonstrate that our default configuration (20\% filtering) yields optimal performance, and the method remains robust, consistently outperforming baselines (CrossScore, PuzzleSim) even under significant geometric noise.}
\begin{adjustbox}{max width=\linewidth}
\begin{tabular}{llcccccccc}
\hline
\multicolumn{1}{c}{\multirow{3}{*}{Method}} & \multicolumn{1}{c}{\multirow{3}{*}{Type}} & \multicolumn{4}{c}{Mip-NeRF 360}                                          & \multicolumn{4}{c}{Tanks and Temples}                                     \\ \cline{3-10} 
\multicolumn{1}{c}{}                        & \multicolumn{1}{c}{}                      & \multicolumn{2}{c}{DINOv2} & \multicolumn{2}{c}{SSIM} & \multicolumn{2}{c}{DINOv2} & \multicolumn{2}{c}{SSIM} \\ \cline{3-10} 
\multicolumn{1}{c}{}                        & \multicolumn{1}{c}{}                      & PLCC          & SRCC        & PLCC        & SRCC       & PLCC          & SRCC        & PLCC        & SRCC        \\ \hline
CrossScore                                  & \multicolumn{1}{c}{-}                     & 0.094         & 0.090       & 0.290       & 0.325      & 0.237         & 0.272       & 0.444       & 0.462       \\
PuzzleSim                                   & \multicolumn{1}{c}{-}                     & 0.304         & 0.327       & 0.128       & 0.124      & 0.351         & 0.369       & 0.348       & 0.347       \\ \hline
% -- Ours_DINOv2 묶음 (5행) --
\multirow{5}{*}{Ours$_{\text{DINOv2}}$}               & (20\% Conf Filtering)                     & \cellcolor{MyRed}0.555         & \cellcolor{MyRed}0.622       & 0.261       & 0.241      & \cellcolor{MyRed}0.573         & \cellcolor{MyRed}0.649       & 0.387       & 0.367       \\
                                            & + 50\% Conf Filtering                     & \cellcolor{MyYellow}0.476         & \cellcolor{MyYellow}0.522       & 0.252       & 0.241      & \cellcolor{MyYellow}0.495         & \cellcolor{MyYellow}0.559       & 0.362       & 0.325       \\
                                            & + No Filtering                            & \cellcolor{MyOrange}0.495         & \cellcolor{MyOrange}0.555       & 0.261       & 0.251      & \cellcolor{MyOrange}0.517         & \cellcolor{MyOrange}0.584       & 0.352       & 0.310       \\
                                            & + 5\% Random Noise on Cam                 & 0.460         & 0.498       & 0.248       & 0.240      & 0.480         & 0.511       & 0.358       & 0.319       \\
                                            & + 10\% Random Noise on Cam                & 0.447         & 0.477       & 0.244       & 0.236      & 0.464         & 0.479       & 0.353       & 0.315       \\ \hline
% -- Ours_SSIM 묶음 (5행) --
\multirow{5}{*}{Ours$_{\text{SSIM}}$}                 & (20\% Conf Filtering)                     & 0.320         & 0.367       & \cellcolor{MyRed}0.535       & \cellcolor{MyRed}0.556      & 0.309         & 0.344       & \cellcolor{MyRed}0.625       & \cellcolor{MyRed}0.642       \\
                                            & + 50\% Conf Filtering                     & 0.301         & 0.348       & \cellcolor{MyYellow}0.514       & \cellcolor{MyYellow}0.534      & 0.294         & 0.326       & 0.607       & \cellcolor{MyYellow}0.624       \\
                                            & + No Filtering                            & 0.304         & 0.349       & \cellcolor{MyOrange}0.520       & \cellcolor{MyOrange}0.542      & 0.301         & 0.334       & \cellcolor{MyYellow}0.609       & \cellcolor{MyOrange}0.626       \\
                                            & + 5\% Random Noise on Cam                 & 0.312         & 0.360       & 0.505       & 0.524      & 0.293         & 0.320       & \cellcolor{MyOrange}0.610       & \cellcolor{MyYellow}0.624       \\
                                            & + 10\% Random Noise on Cam                & 0.312         & 0.360       & 0.504       & 0.523      & 0.292         & 0.318       & \cellcolor{MyYellow}0.609       & \cellcolor{MyYellow}0.624       \\ \hline
\end{tabular}
\end{adjustbox}
\label{T6_geometric}
\end{table*}

\begin{table}[t]
\centering
\caption{Comparison of FR-IQA metrics as guidance signals for Quality-Aware 3DGS training. We evaluate the 3DGS modeling quality (PSNR, SSIM, LPIPS) when guiding the optimization using different IQA targets (PSNR, SSIM, LPIPS, and DINOv2). The results demonstrate that DINOv2 feature similarity consistently outperforms traditional metrics, even surpassing methods that directly optimize for the target metric itself, thereby justifying its selection as our primary prediction target.}
\begin{adjustbox}{max width=\linewidth}
\begin{tabular}{ccccccc}
\hline
                     & \multicolumn{3}{c}{Tanks and Temples}         & \multicolumn{3}{c}{Mip-NeRF 360}               \\ \hline
IQA method           & PSNR           & SSIM            & LPIPS           & PSNR           & SSIM            & LPIPS           \\ \hline
PSNR                 & 7.09           & 0.435           & 0.575           & 7.33           & 0.371           & 0.574           \\
SSIM                 & \underline{14.11}    & \underline{0.525}     & \underline{0.482}     & \underline{15.71}    & \underline{0.515}     & \underline{0.466}     \\
LPIPS                & 13.21          & 0.524           & 0.480           & 14.72          & 0.502           & 0.472           \\
DINOv2-SIM & \textbf{16.05} & \textbf{0.548}  & \textbf{0.465}  & \textbf{18.29} & \textbf{0.526}  & \textbf{0.453}  \\ \hline
\end{tabular}
\end{adjustbox}
\label{T7_feature}
\end{table}

\begin{table*}[t]
\centering
\caption{Ablation on the masking threshold $\tau$ for 3DGS training. We evaluate the impact of the pixel retention rate $\tau$ on reconstruction quality. We compare aggressive ($\tau=30$), default ($\tau=50$), and lenient ($\tau=70$) filtering strategies on the Mip-NeRF 360 and Tanks and Temples datasets. The results show that $\tau=50$ achieves the best performance across datasets, validating it as a robust heuristic that balances noise removal with data retention. Red, orange, and yellow cells denote the 1st, 2nd, and 3rd best methods per column. (excluding FR settings †)}
\begin{adjustbox}{max width=\linewidth}
\begin{tabular}{llcccccc|cccccc|cccccc}
\hline
                                              &                  & \multicolumn{6}{c|}{\(\tau=30\)}                                                                                                                                                      & \multicolumn{6}{c|}{\(\tau=50\)}                                                                                                                                                      & \multicolumn{6}{c}{\(\tau=70\)}                                                                                                                                                      \\ \hline
                                              &                  & \multicolumn{3}{c}{Mip-NeRF 360}                                                               & \multicolumn{3}{c|}{Tanks and Temples}                                                          & \multicolumn{3}{c}{Mip-NeRF 360}                                                               & \multicolumn{3}{c|}{Tanks and Temples}                                                          & \multicolumn{3}{c}{Mip-NeRF 360}                                                               & \multicolumn{3}{c}{Tanks and Temples}                                                          \\ \hline
IQA-Guided                                    & Method           & PSNR                           & SSIM                          & LPIPS                         & PSNR                           & SSIM                          & LPIPS                         & PSNR                           & SSIM                          & LPIPS                         & PSNR                           & SSIM                          & LPIPS                         & PSNR                           & SSIM                          & LPIPS                         & PSNR                           & SSIM                          & LPIPS                         \\ \hline
\multicolumn{1}{c}{}                          & Vanilla 3DGS     & 16.078                         & 0.461                         & \cellcolor{MyOrange}0.415 & 15.298                         & 0.509                         & \cellcolor[HTML]{FFA0A0}0.406 & 16.078                         & 0.461                         & \cellcolor{MyOrange}0.415 & 15.298                         & 0.509                         & \cellcolor{MyOrange}0.406 & 16.078                         & 0.461                         & \cellcolor[HTML]{FFA0A0}0.415 & 15.298                         & 0.509                         & \cellcolor{MyOrange}0.406 \\
\multicolumn{1}{c}{\multirow{-2}{*}{w/o IQA}} & ViewCrafter      & 16.179                         & \cellcolor{MyOrange}0.474 & 0.452                         & \cellcolor{MyYellow}15.773 & \cellcolor{MyYellow}0.523 & 0.455                         & 16.179                         & 0.474                         & 0.452                         & 15.773                         & 0.523                         & 0.455                         & 16.179                         & 0.474                         & 0.452                         & 15.773                         & 0.523                         & 0.455                         \\ \hline
\hline
                                              & SSIM$^\dagger$             & 16.837                         & 0.494                         & 0.413                         & 16.331                         & 0.551                         & 0.397                         & 16.676                         & 0.487                         & 0.421                         & 16.228                         & 0.556                         & 0.399                         & 16.779                         & 0.491                         & 0.425                         & 16.405                         & 0.557                         & 0.407                         \\
\multirow{-2}{*}{w/ FR-IQA}                   & DINOv2$^\dagger$           & 16.892                         & 0.494                         & 0.400                         & 16.401                         & 0.551                         & 0.392                         & 17.178                         & 0.498                         & 0.399                         & 16.777                         & 0.562                         & 0.384                         & 17.209                         & 0.497                         & 0.412                         & 16.784                         & 0.560                         & 0.391                         \\ \hline
                                              & PaQ-2-PiQ        & 16.148                         & 0.456                         & 0.432                         & 15.345                         & 0.511                         & 0.435                         & 16.298                         & 0.472                         & 0.425                         & 15.769                         & 0.534                         & 0.421                         & 16.370                         & 0.477                         & 0.430                         & 16.137                         & 0.546                         & 0.414                         \\
\multirow{-2}{*}{w/ NR-IQA}                   & PIQE             & 15.858                         & 0.462                         & 0.450                         & 15.275                         & 0.521                         & 0.443                         & 16.313                         & 0.479                         & 0.440                         & 15.671                         & 0.534                         & 0.433                         & 16.426                         & 0.478                         & 0.441                         & 15.936                         & 0.543                         & 0.427                         \\ \hline
                                              & CrossScore       & 16.036                         & 0.469                         & 0.441                         & 15.196                         & 0.515                         & 0.440                         & 16.312                         & 0.476                         & 0.431                         & 15.856                         & 0.537                         & 0.427                         & 16.463                         & \cellcolor{MyYellow}0.480 & 0.442                         & \cellcolor{MyYellow}16.195 & \cellcolor{MyYellow}0.547 & 0.424                         \\
                                              & PuzzleSim        & \cellcolor{MyYellow}16.239 & \cellcolor{MyYellow}0.473 & \cellcolor[HTML]{FFA0A0}0.411 & 15.645                         & \cellcolor{MyOrange}0.527 & \cellcolor{MyYellow}0.414 & \cellcolor{MyYellow}16.349 & \cellcolor{MyYellow}0.482 & \cellcolor{MyYellow}0.423 & \cellcolor{MyYellow}15.937 & \cellcolor{MyYellow}0.541 & \cellcolor{MyOrange}0.406 & \cellcolor{MyYellow}16.469 & 0.479                         & \cellcolor{MyOrange}0.421 & 16.104                         & 0.546                         & \cellcolor{MyOrange}0.406 \\
                                              & Ours$_{\text{SSIM}}$   & \cellcolor{MyOrange}16.319 & \cellcolor{MyOrange}0.474 & 0.437                         & \cellcolor{MyOrange}15.914 & \cellcolor[HTML]{FFA0A0}0.540 & \cellcolor{MyOrange}0.410 & \cellcolor{MyOrange}16.371 & \cellcolor{MyOrange}0.485 & 0.427                         & \cellcolor{MyOrange}16.143 & \cellcolor{MyOrange}0.548 & \cellcolor{MyYellow}0.407 & \cellcolor{MyOrange}16.512 & \cellcolor{MyOrange}0.482 & 0.437                         & \cellcolor{MyOrange}16.240 & \cellcolor{MyOrange}0.551 & \cellcolor{MyYellow}0.416 \\
\multirow{-4}{*}{w/ CR-IQA}                   & Ours$_{\text{DINOv2}}$ & \cellcolor[HTML]{FFA0A0}16.529 & \cellcolor[HTML]{FFA0A0}0.482 & \cellcolor{MyYellow}0.417 & \cellcolor[HTML]{FFA0A0}15.981 & \cellcolor[HTML]{FFA0A0}0.540 & \cellcolor[HTML]{FFA0A0}0.406 & \cellcolor[HTML]{FFA0A0}16.756 & \cellcolor[HTML]{FFA0A0}0.493 & \cellcolor[HTML]{FFA0A0}0.414 & \cellcolor[HTML]{FFA0A0}16.238 & \cellcolor[HTML]{FFA0A0}0.551 & \cellcolor[HTML]{FFA0A0}0.403 & \cellcolor[HTML]{FFA0A0}16.736 & \cellcolor[HTML]{FFA0A0}0.489 & \cellcolor{MyYellow}0.424 & \cellcolor[HTML]{FFA0A0}16.370 & \cellcolor[HTML]{FFA0A0}0.554 & \cellcolor[HTML]{FFA0A0}0.405 \\ \hline
\end{tabular}
\end{adjustbox}
\begin{tablenotes}
\footnotesize
\item $\dagger$ Metrics require a same-pose GT image.
\end{tablenotes}
\label{T8_tau}
\end{table*}

\subsection{Ablation Study on Loss Components}

In this section, we evaluate the contribution of the individual loss terms defined in the training objective (Eq.~(5) in the main manuscript). Our full objective function combines a pixel-wise reconstruction loss ($\mathcal{L}_1$) with two distribution-aware losses: the Jensen-Shannon Divergence loss ($\mathcal{L}_{\text{JSD}}$) and the Pearson Linear Correlation Coefficient loss ($\mathcal{L}_{\text{PLCC}}$). To isolate the impact of these auxiliary terms, we trained variants of our model by removing them one at a time.

Table~\ref{T5_loss} presents the performance comparison on the Mip-NeRF 360 and Tanks and Temples datasets. The results demonstrate that $\mathcal{L}_{\text{JSD}}$ and $\mathcal{L}_{\text{PLCC}}$ are not merely supplementary but are fundamental to the learning process.

As shown in the table, removing either the JSD loss (``w/o $\mathcal{L}_{\text{JSD}}$'') or the PLCC loss (``w/o $\mathcal{L}_{\text{PLCC}}$'') leads to a catastrophic performance drop, resulting in negative correlation values across all metrics and datasets. A negative correlation implies that the model's predictions are inversely related to the GT, indicating a complete failure to learn the correct quality ranking.

\begin{table}[t]
\centering
\caption{Ablation study comparing binary masking and soft masking strategies. We evaluate the robustness of our framework by comparing the default binary masking approach against a continuous soft weighting strategy.}
\begin{adjustbox}{max width=\linewidth}
\begin{tabular}{llcccccc}
\hline
                             &              & \multicolumn{3}{c}{Mip-NeRF 360} & \multicolumn{3}{c}{Tanks and Temples} \\ \hline
Mask Type                    & Method       & PSNR      & SSIM      & LPIPS    & PSNR        & SSIM       & LPIPS      \\ \hline
\multirow{2}{*}{Binary Mask} & Ours$_{\text{SSIM}}$   & 16.37     & 0.485     & 0.427    & 16.14       & 0.548      & 0.407      \\
                             & Ours$_{\text{DINOv2}}$ & 16.76     & \textbf{0.493}     & \textbf{0.414}    & 16.24       & 0.551      & \textbf{0.403}      \\ \hline
\multirow{2}{*}{Soft Mask}   & Ours$_{\text{SSIM}}$   & 16.61     & 0.485     & 0.437    & 16.34       & \textbf{0.553}      & 0.418      \\
                             & Ours$_{\text{DINOv2}}$ & \textbf{16.78}     & 0.488     & 0.426    & \textbf{16.44}       & \textbf{0.553}      & 0.405      \\ \hline
\end{tabular}
\end{adjustbox}
\label{T9_binary_soft}
\end{table}

In contrast, the ``Full Model'' achieves strong positive correlations (e.g., PLCC $> 0.55$). This sharp contrast suggests that the pixel-wise loss alone is insufficient for this task. The combination of $\mathcal{L}_{\text{JSD}}$ (which aligns score distributions) and $\mathcal{L}_{\text{PLCC}}$ (which enforces linear relationship) provides the necessary constraints to stabilize training and guide the model toward perceptually meaningful quality predictions.

\subsection{Geometric Robustness Analysis}
\label{geometric}

In this section, we investigate the sensitivity of the PR-IQA framework to geometric imperfections, specifically focusing on point cloud quality and camera pose accuracy. As detailed in our methodology (Sect. 3.3 of the main manuscript), our approach generates a partial quality map by warping features from the reference image to the query view using VGGT~\cite{Wang_2025_CVPR}. This process relies on estimating 3D points via stereo correspondences and reprojecting them for feature alignment. To mitigate artifacts arising from unreliable correspondences, our default configuration filters out 3D points falling within the bottom 20\% of confidence scores, utilizing only the remaining high-confidence points for warping. To evaluate the robustness of this design, we conducted experiments varying this filtering threshold and introducing synthetic noise to the estimated camera poses.

Table~\ref{T6_geometric} summarizes the performance of our method under these varying geometric conditions. A broad analysis reveals that our proposed methods ($\text{Ours}_{\text{DINOv2}}$ and $\text{Ours}_{\text{SSIM}}$) consistently achieve significantly higher PLCC and SRCC correlations compared to baselines like CrossScore and PuzzleSim across both Mip-NeRF 360 and Tanks and Temples datasets. This empirically validates the effectiveness of our geometry-guided feature matching approach.

\paragraph{Impact of Point Cloud Filtering.}
We analyzed how the density and reliability of the geometric input affect performance by adjusting the VGGT depth confidence filter. As shown in Table~\ref{T6_geometric}, the default setting, removing the bottom 20\% of low-confidence points, yields optimal performance. This threshold strikes a critical balance: it effectively eliminates high-variance noise (e.g., sky regions or inaccurate depths) while preserving sufficient scene context essential for matching. Conversely, performance degrades under the ``No Filtering'' setting due to the inclusion of geometric outliers, as well as under the stricter ``+50\% Conf Filtering'' setting, where the excessive removal of points leads to a loss of valuable visual information.

\paragraph{Robustness to Camera Pose Noise.}
To evaluate resilience against inaccurate camera poses, a common challenge in real-world sparse-view reconstruction, we introduced Gaussian noise to both intrinsic and extrinsic parameters. We defined two noise levels:

\begin{itemize}
    \item \textbf{5\% Noise Level:} Perturbations included rotation by approximately $5^\circ$, translation by 5\% of the original magnitude, focal length by 5\%, and principal point shifts by 5\% of image dimensions.
    \item \textbf{10\% Noise Level:} These perturbations were doubled (e.g., approximately $10^\circ$ rotation).
\end{itemize}

As expected, the performance exhibits a gradual decline as noise levels increase (see Table~\ref{T6_geometric}). However, a crucial finding is that even under significant perturbations (10\% noise), our method maintains competitive scores that continue to surpass the baseline methods (CrossScore and PuzzleSim). This confirms that the PR-IQA framework is not only effective under ideal conditions but also practically robust to the geometric errors frequently encountered in sparse-view scenarios.

\begin{table*}
\centering
\setlength{\tabcolsep}{2.5pt}
\caption{Low-overlap evaluation. We evaluate robustness by grouping image pairs from the original dataset by overlap ratio.}
\label{tab:low_overlap}
\begin{adjustbox}{max width=0.85\linewidth}
\begin{threeparttable}
\begin{tabular}{lcccccccccccccccc}
\toprule
\multirow{3}{*}{IQA Method} &
\multicolumn{4}{c}{25\% (81)} &
\multicolumn{4}{c}{20\% (52)} &
\multicolumn{4}{c}{10\% (17)} &
\multicolumn{4}{c}{5\% (9)} \\
\cmidrule(lr){2-5}\cmidrule(lr){6-9}\cmidrule(lr){10-13}\cmidrule(lr){14-17}
& \multicolumn{2}{c}{DINOv2} & \multicolumn{2}{c}{SSIM}
& \multicolumn{2}{c}{DINOv2} & \multicolumn{2}{c}{SSIM}
& \multicolumn{2}{c}{DINOv2} & \multicolumn{2}{c}{SSIM}
& \multicolumn{2}{c}{DINOv2} & \multicolumn{2}{c}{SSIM} \\
\cmidrule(lr){2-3}\cmidrule(lr){4-5}
\cmidrule(lr){6-7}\cmidrule(lr){8-9}
\cmidrule(lr){10-11}\cmidrule(lr){12-13}
\cmidrule(lr){14-15}\cmidrule(lr){16-17}
& PLCC & SRCC & PLCC & SRCC
& PLCC & SRCC & PLCC & SRCC
& PLCC & SRCC & PLCC & SRCC
& PLCC & SRCC & PLCC & SRCC \\
\midrule
CrossScore
& 0.137 & 0.118 & \cellcolor{MyYellow}{0.153} & \cellcolor{MyYellow}{0.139}
& \cellcolor{MyYellow}{0.189} & 0.162 & \cellcolor{MyYellow}{0.106} & \cellcolor{MyYellow}{0.074}
& \cellcolor{MyYellow}{0.223} & \cellcolor{MyYellow}{0.194} & \cellcolor{MyYellow}{0.141} & \cellcolor{MyYellow}{0.079}
& \cellcolor{MyYellow}{0.250} & \cellcolor{MyYellow}{0.217} & \cellcolor{MyYellow}{0.131} & \cellcolor{MyYellow}{0.080} \\

PuzzleSim
& \cellcolor{MyYellow}{0.178} & \cellcolor{MyYellow}{0.211} & 0.013 & 0.004
& 0.120 & \cellcolor{MyYellow}{0.173} & 0.041 & 0.030
& 0.041 & 0.081 & 0.024 & 0.006
& -0.007 & 0.027 & 0.058 & 0.041 \\

Ours$_{\text{DINOv2}}$
& \cellcolor{MyRed}{0.469} & \cellcolor{MyRed}{0.502} & \cellcolor{MyOrange}{0.374} & \cellcolor{MyOrange}{0.366}
& \cellcolor{MyRed}{0.485} & \cellcolor{MyRed}{0.503} & \cellcolor{MyOrange}{0.396} & \cellcolor{MyOrange}{0.382}
& \cellcolor{MyRed}{0.501} & \cellcolor{MyRed}{0.511} & \cellcolor{MyOrange}{0.383} & \cellcolor{MyOrange}{0.370}
& \cellcolor{MyRed}{0.486} & \cellcolor{MyRed}{0.484} & \cellcolor{MyOrange}{0.386} & \cellcolor{MyOrange}{0.369} \\

Ours$_{\text{SSIM}}$
& \cellcolor{MyOrange}{0.278} & \cellcolor{MyOrange}{0.295} & \cellcolor{MyRed}{0.463} & \cellcolor{MyRed}{0.482}
& \cellcolor{MyOrange}{0.331} & \cellcolor{MyOrange}{0.311} & \cellcolor{MyRed}{0.462} & \cellcolor{MyRed}{0.477}
& \cellcolor{MyOrange}{0.384} & \cellcolor{MyOrange}{0.345} & \cellcolor{MyRed}{0.434} & \cellcolor{MyRed}{0.414}
& \cellcolor{MyOrange}{0.365} & \cellcolor{MyOrange}{0.300} & \cellcolor{MyRed}{0.409} & \cellcolor{MyRed}{0.418} \\

\bottomrule
\end{tabular}

\begin{tablenotes}[flushleft]
\normalsize
\item \textbf{Note.} \%: overlap ratio; (): \# of images.
\end{tablenotes}
\end{threeparttable}
\end{adjustbox}
\end{table*}

\begin{figure*}
    \centering
    \includegraphics[width=0.9\linewidth]{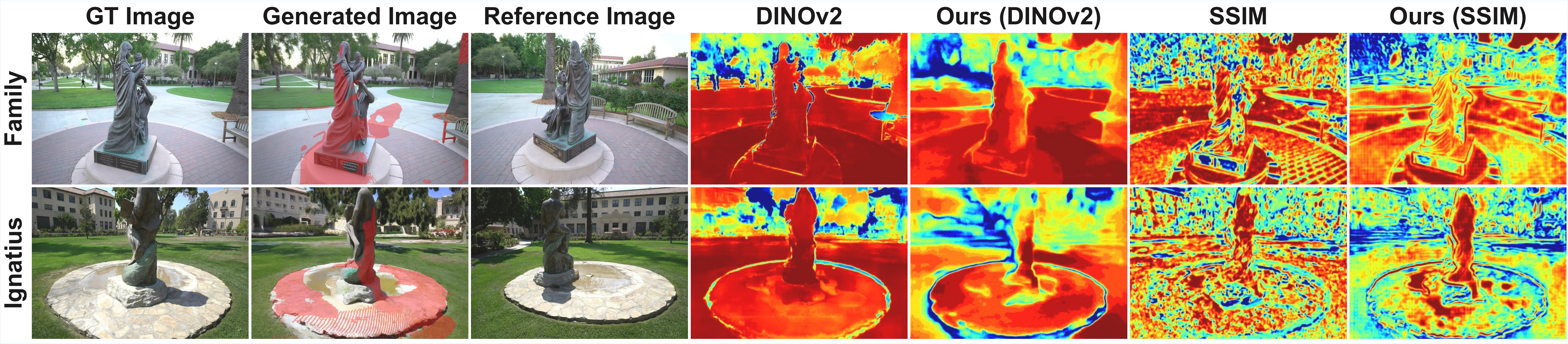}
    \caption{Low-overlap qualitative results. Red region in the generated image shows overlaps of 16\% {\footnotesize (Family)}  and 22\% {\footnotesize (Ignatius)}. }
    \label{fig:low_overlap}
\end{figure*}

\subsection{Low-Overlap Robustness Analysis}
\label{sec:low_overlap}

To examine robustness under limited visual correspondence, we re-evaluated the test set by regrouping image pairs according to their overlap ratio. As shown in Table~\ref{tab:low_overlap}, the proposed PR-IQA remains stable even as the overlap becomes progressively smaller, while competing methods tend to degrade more noticeably under the same condition. This result indicates that our method does not rely solely on directly shared regions between the generated and reference images, but instead learns transferable quality cues that remain meaningful when only partial correspondence is available.

Fig.~\ref{fig:low_overlap} further illustrates this behavior in challenging low-overlap examples. Even when the common visible region is very limited, our method produces quality maps that better preserve the perceptually important structure and object-level consistency than direct full-reference targets. In particular, the propagated responses remain coherent beyond the overlapping area, supporting reliable quality estimation in non-overlapping regions. These observations confirm that PR-IQA effectively extends local reference evidence into the unseen area and remains robust even in near-zero-overlap cases.

\begin{table}
\caption{FPR@Top-$X\%$ measures how often pixels in the top $X\%$ of scores within non-overlapping regions are falsely rated as high quality on Tanks and Temples.}
\label{tab:false_positive}
\centering
\setlength{\tabcolsep}{3pt}
\begin{adjustbox}{max width=\linewidth}
\begin{tabular}{lccccc}
\toprule
Method & FPR@50\% & FPR@40\% & FPR@30\% & FPR@20\% & FPR@10\% \\
\midrule
CrossScore            & 0.380 & 0.300 & 0.240 & 0.183 & 0.105 \\
PuzzleSim             & 0.328 & 0.273 & 0.222 & 0.162 & 0.093 \\
Ours$_{\text{DINOv2}}$ & \textbf{0.306} & \textbf{0.236} & \textbf{0.183} & \textbf{0.137} & \textbf{0.082} \\
\bottomrule
\end{tabular}
\end{adjustbox}
\end{table}

\begin{figure*}
    \centering
    \includegraphics[width=0.9\linewidth]{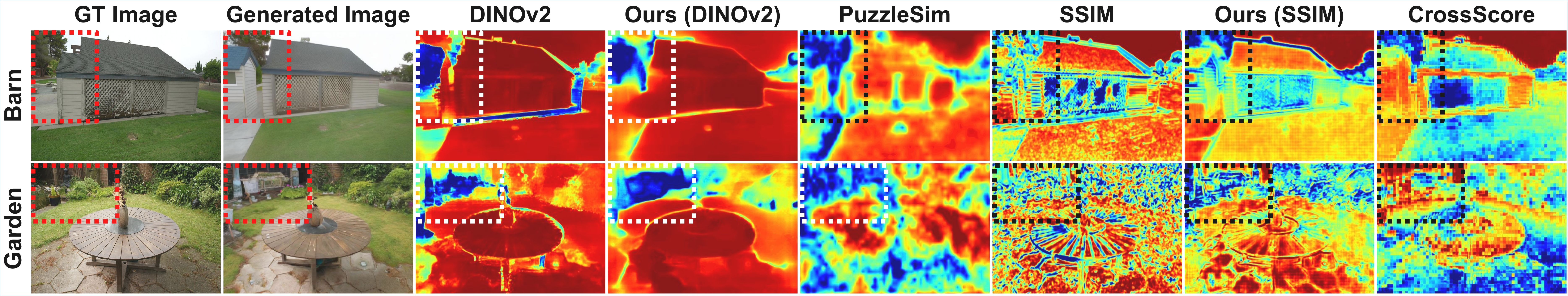}
    \caption{Quality estimation results on hallucinated non-overlapping regions (boxed) from the Barn and Garden scenes. The dashed boxes highlight unsupported areas that are visible in the generated image but not reliably matched to the reference view.}
    \label{fig:false_positive}
\end{figure*}

\subsection{False Positive Analysis in Non-Overlapping Regions}
\label{sec:false_positive}

To further analyze reliability in unseen areas, we measured the \textit{False Positive Rate} (FPR@Top-$X\%$) specifically within non-overlapping regions, where hallucinated content is most likely to appear. As summarized in Table~\ref{tab:false_positive}, Ours$_{\text{DINOv2}}$ consistently achieves the lowest false positive rate across all evaluation thresholds. This indicates that the proposed PR-IQA is less prone to incorrectly assigning high-quality scores to regions that are not supported by reference evidence, demonstrating stronger conservativeness and robustness in ambiguous areas.

Fig.~\ref{fig:false_positive} provides qualitative examples of this behavior on hallucinated objects and structures. Compared with existing methods, our predictions suppress spuriously high responses in boxed non-overlapping regions while preserving meaningful quality patterns in the valid area. In contrast, competing approaches more often produce overly confident activations on unsupported content. These results confirm that PR-IQA avoids false positives on hallucinated content.
% These results confirm that PR-IQA not only propagates useful quality cues beyond overlapping regions, but also better avoids false positives on hallucinated content.

\section{More Ablation Studies on 3DGS}

\subsection{Effectiveness of DINOv2 Feature Similarity}
We validate the rationale behind selecting DINOv2 feature similarity (i.e., DINOv2-SIM) as our primary optimization target by comparing its effectiveness against standard FR-IQA metrics: PSNR, SSIM, and LPIPS. To ensure a fair comparison, we integrated these metrics into the ``Quality-Aware 3DGS Training'' pipeline (described in Section~4 of the main manuscript) as alternative guidance signals. For consistency, all quality maps were normalized to the range $[0, 1]$ via min-max scaling, where higher values denote better quality.

As detailed in Table~\ref{T7_feature}, the 3DGS reconstruction guided by DINOv2-based quality maps consistently yields superior performance across all evaluation metrics on both the Tanks and Temples and Mip-NeRF 360 datasets. A remarkable finding is that utilizing DINOv2 similarity as a training guide results in higher final PSNR and SSIM scores than using those specific metrics themselves as guidance targets.

This superiority stems from the inherent limitations of conventional metrics in the context of diffusion-based synthesis. Pixel-wise metrics like PSNR tend to unduly penalize regions that possess valid geometric structures but exhibit minor color shifts or lighting variations, thereby discarding potentially useful supervision signals. Similarly, SSIM and LPIPS often struggle to reliably distinguish between fine geometric details and artifacts in generated views. In contrast, our DINOv2-based approach prioritizes high-level semantic and geometric alignment. It effectively identifies and utilizes structurally consistent regions while remaining robust to benign photometric discrepancies, making it significantly more suitable for supervising 3D reconstruction from diffusion-generated imagery.

\subsection{Impact of Masking Threshold \texorpdfstring{$\tau$}{tau}}

In this section, we provide a detailed ablation study to validate our choice of the masking threshold $\tau$, which was set to a heuristic value of $50$ in the main manuscript. In our framework, $\tau$ represents the retention rate, the percentage of pixels with the highest predicted quality scores that are used for 3DGS optimization. We determine a global quality threshold $Q_{\text{thresh}}$ corresponding to the $(100-\tau)$-th percentile of the score distribution; pixels exceeding this value are included in the training mask. Thus, a lower $\tau$ (e.g., $\tau = 30$) implies an aggressive filtering strategy that retains only the top $30\%$ of pixels, whereas a higher $\tau$ (e.g., $\tau = 70$) is more lenient.

Table~\ref{T8_tau} presents the reconstruction performance across varying thresholds ($\tau \in \{30, 50, 70\}$). The aggressive strategy ($\tau=30$) consistently yields the lowest performance across both datasets. This indicates that while removing low-quality regions is essential, discarding $70\%$ of the generated data eliminates too much valid supervision signal, thereby hindering the geometry convergence and degrading the final reconstruction quality.

Performance peaks between $\tau=50$ and $\tau=70$. For $\text{Ours}_{\text{DINOv2}}$, the heuristic $\tau=50$ achieves the best PSNR ($16.756$) on the Mip-NeRF 360 dataset, outperforming both the stricter ($\tau=30$) and looser ($\tau=70$) settings. On the Tanks and Temples dataset, while $\tau=70$ yields a marginal improvement, the performance at $\tau=50$ remains highly competitive and robust.

This study confirms that $\tau=50$ serves as an effective and robust heuristic across diverse scenes. It strikes a critical balance: it is strict enough to filter out significant artifacts and inconsistencies, yet lenient enough to preserve a sufficient density of high-confidence pseudo-ground-truth pixels for accurate 3D reconstruction.

\begin{table}[t]
    \caption{Computational cost analysis. We report the averaged runtime (seconds) and memory usage (MB) for individual components of the PR-IQA pipeline and the 3DGS optimization process.}
    \centering
    \small
    \begin{tabular}{l l c c}
        \hline
        Method & Stage & Runtime (s) & Memory (MB) \\
        \hline
        \multirow{3}{*}{PR-IQA} 
            & Feature Ext.  & 0.303 & 5509.250 \\
            & VGGT         & 0.207 & 2448.317 \\
            & Inference    & 0.510 & 530.950 \\
        \hline
        3DGS  & - & 25.210 & 749.780 \\
        \hline
    \end{tabular}
    \label{T10_computation}
\end{table}

\subsection{Soft vs. Binary Masking Strategies}

In our primary manuscript, we employ a binary masking strategy that strictly includes or excludes pixels based on a confidence threshold. In this section, we conduct an ablation study to evaluate an alternative ``soft weighting'' strategy. Instead of a hard binary selection ($0$ or $1$), this approach utilizes the predicted continuous quality score directly as a pixel-wise loss weight (ranging from $0$ to $1$) during 3DGS optimization. This allows the influence of each pixel to be modulated gradually by its estimated quality.

\paragraph{Mathematical Formulation.}
Let $\mathcal{L}_{\text{base}}(p)$ denote the standard photometric loss (e.g., $\mathcal{L}_1$ or D-SSIM) for a pixel $p$ during 3DGS training.

\begin{itemize}
    \item \textbf{Binary Masking:} We define a binary mask $M(p)$ based on the quality threshold $Q_{\tau}$ derived from the percentile $\tau$:
    \begin{equation}
        M(p) = \mathbf{1}(Q(p) \ge Q_{\tau}) = 
        \begin{cases} 
        1 & \text{if } Q(p) \ge Q_{\tau} \\
        0 & \text{otherwise}
        \end{cases}.
    \end{equation}
    The final loss function is given by:
    \begin{equation}
        \mathcal{L}_{\text{binary}} = \sum_{p \in \mathcal{P}} M(p) \cdot \mathcal{L}_{\text{base}}(p).
    \end{equation}
    
    \item \textbf{Soft Weighting:} We directly use the normalized predicted quality score $Q(p) \in [0, 1]$ as a weighting factor $W(p)$:
    \begin{equation}
        W(p) = Q(p).
    \end{equation}
    The weighted loss function becomes:
    \begin{equation}
        \mathcal{L}_{\text{soft}} = \sum_{p \in \mathcal{P}} W(p) \cdot \mathcal{L}_{\text{base}}(p).
    \end{equation}
\end{itemize}

Table~\ref{T9_binary_soft} compares the reconstruction performance of our method under both masking regimes. The quantitative results indicate that both strategies yield highly similar performance metrics across the Mip-NeRF 360 and Tanks and Temples datasets. For instance, while binary masking achieves a slightly better LPIPS score on Mip-NeRF 360, soft masking yields a marginally higher PSNR. Overall, the performance differences are negligible, suggesting that both approaches effectively guide the optimization process.

These findings demonstrate the inherent robustness of the PR-IQA framework. The fact that the optimization remains stable and high-performing under both hard-thresholding and continuous-weighting schemes confirms that our predicted quality maps provide reliable supervision signals regardless of the specific masking implementation. This flexibility suggests that practitioners can select either approach, prioritizing the interpretability of binary masks or the differentiability of soft weights, without compromising reconstruction quality.

\subsection{Computational Analysis}
In this section, we evaluate the computational efficiency of the proposed PR-IQA framework. Table~\ref{T10_computation} details the runtime and memory usage for each stage of the pipeline: feature extraction, VGGT-based warping, and quality inference, measured on a single-image basis. A key advantage of our design is that the pipeline internally resizes all inputs to a fixed resolution, ensuring that these computational metrics remain invariant regardless of the original input image resolution.

To provide context for these costs, we compare them against the resource consumption of the standard 3DGS optimization process. This comparison was conducted on the `Barn' scene from the Tanks and Temples dataset, initialized with 28,290 points.

As shown in Table~\ref{T10_computation}, the total runtime for the PR-IQA pipeline is approximately $1.02$ seconds per image (summing feature extraction, VGGT, and inference). In contrast, the 3DGS optimization for the corresponding scene requires $25.21$ seconds. This indicates that the additional computational overhead introduced by our quality assessment module is negligible, making it a highly practical addition to the reconstruction pipeline without causing significant bottlenecks.

\section{More Qualitative Results}

\subsection{More Qualitative Results for Quality Map}
\begin{figure*}[p]
    \centering
    
    \begin{minipage}[t]{0.98\linewidth}
        \centering
        \includegraphics[width=\linewidth]{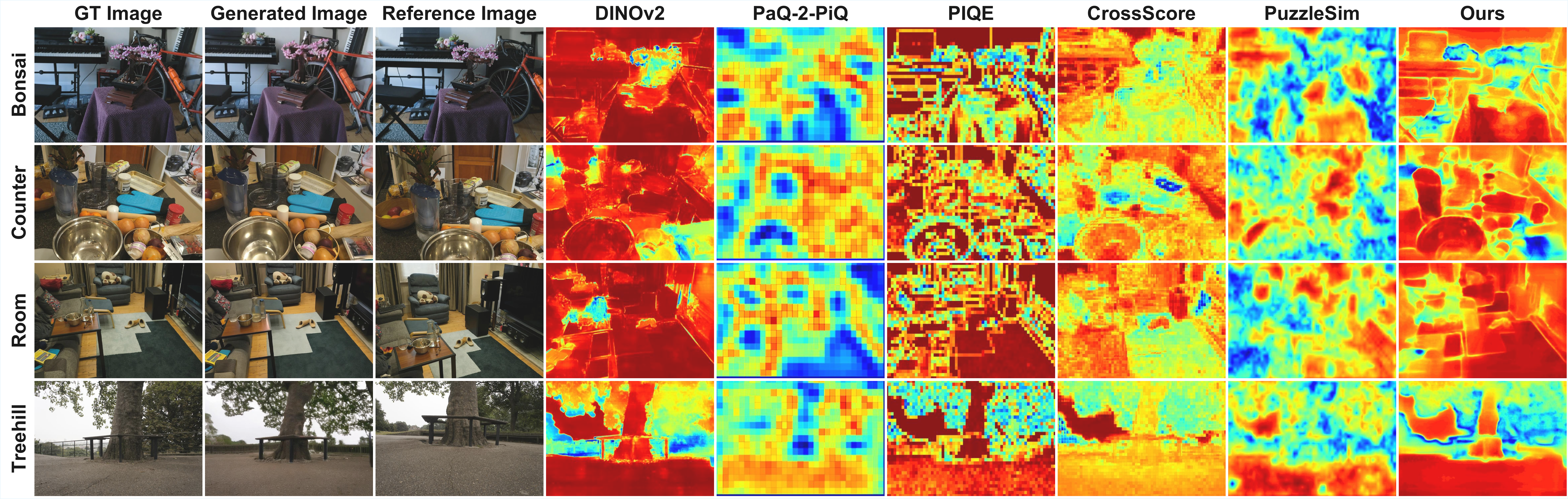}
        \captionof{figure}{Additional quality map comparisons on Mip-NeRF 360 dataset (DINOv2-SIM target). Our method produces quality maps closely aligned with ground-truth DINOv2-SIM.}
        \label{F4_dino_mipnerf}
    \end{minipage}
    
    \vspace{5mm}
    
    \begin{minipage}[t]{0.98\linewidth}
        \centering
        \includegraphics[width=\linewidth]{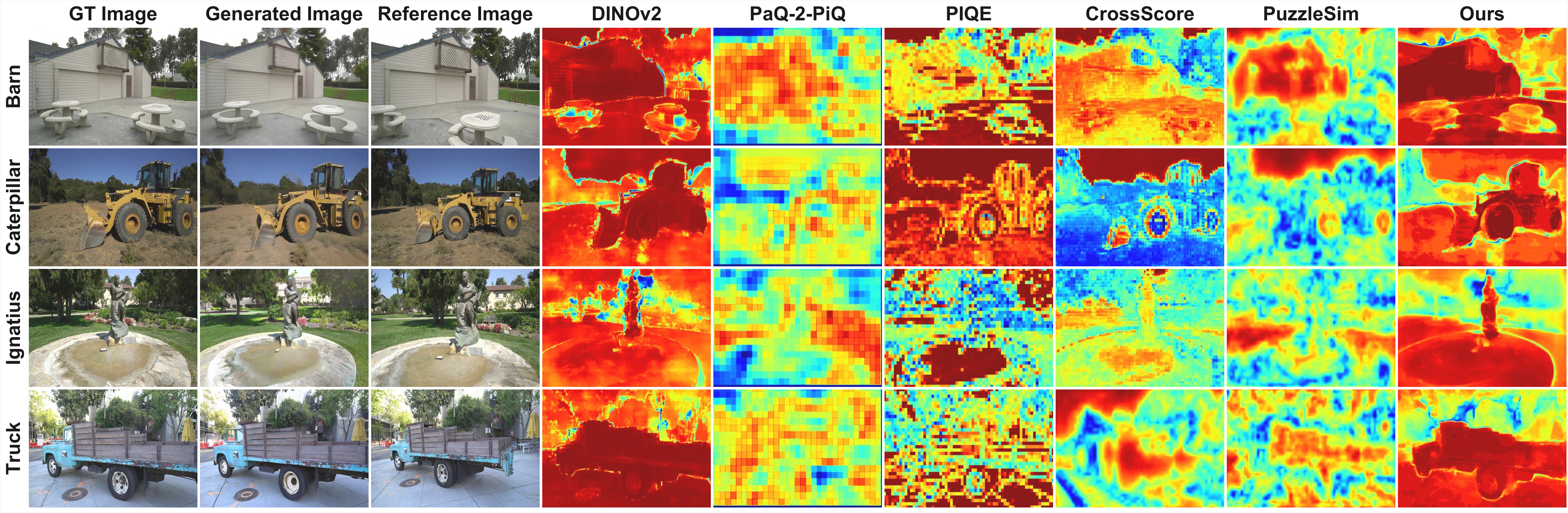}
        \captionof{figure}{Additional quality map comparisons on Tanks and Temples dataset (DINOv2-SIM target). Our PR-IQA consistently estimates quality across complex outdoor scenes.}
        \label{F5_dino_tandt}
    \end{minipage}
    
    \vspace{5mm}
    
    \begin{minipage}[t]{0.98\linewidth}
        \centering
        \includegraphics[width=\linewidth]{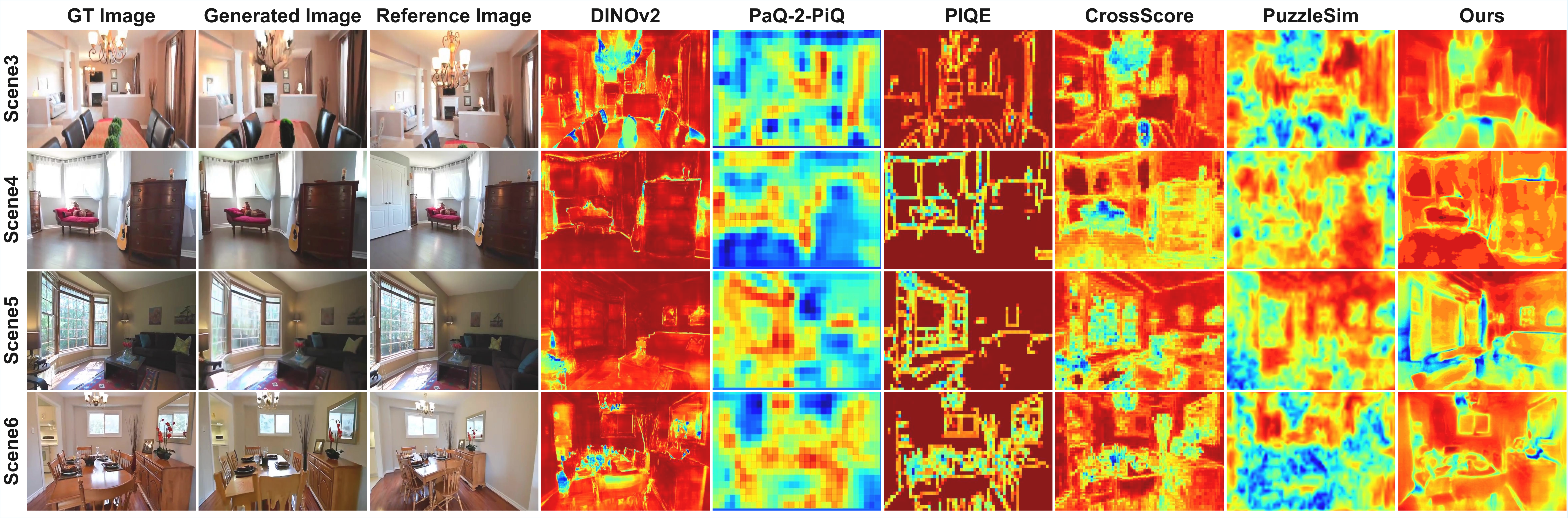}
        \captionof{figure}{Additional quality map comparisons on RealEstate10K dataset (DINOv2-SIM target). Our method demonstrates robust performance on real estate scenes.}
        \label{F6_dino_re10k}
    \end{minipage}
\end{figure*}

We provide extensive qualitative comparisons on scenes not featured in the main manuscript. Figs.~\ref{F4_dino_mipnerf}, \ref{F5_dino_tandt}, and \ref{F6_dino_re10k} illustrate results across the Mip-NeRF 360, Tanks and Temples, and RealEstate10K datasets, respectively. As shown in these figures, our PR-IQA generates quality maps that exhibit high fidelity to the GT DINOv2-SIM, accurately capturing fine-grained variations and sharp boundaries. In contrast, NR-IQA methods often struggle to provide meaningful estimates, while CR-IQA baselines tend to suffer from blocky artifacts, particularly in non-overlapping regions. Our method overcomes these limitations by effectively propagating quality information globally, resulting in smooth and accurate dense quality maps.

\subsection{More Qualitative Results for SSIM Map}
\begin{figure*}[p]
    \centering
    
    \begin{minipage}[t]{0.98\linewidth}
        \centering
        \includegraphics[width=\linewidth]{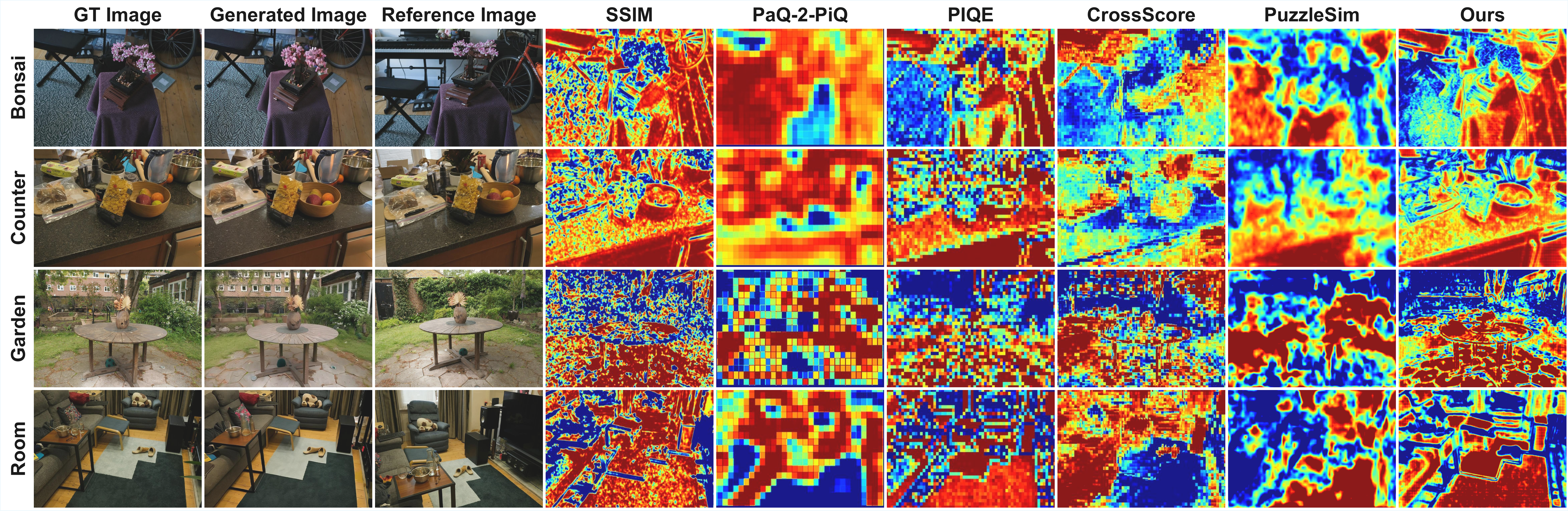}
        \captionof{figure}{Additional quality map comparisons on Mip-NeRF 360 dataset (SSIM target). Ours$_{\text{SSIM}}$ variant effectively predicts SSIM maps, outperforming CrossScore across indoor scenes with various textures and structures.}
        \label{F7_ssim_mipnerf}
    \end{minipage}
    
    \vspace{5mm}
    
    \begin{minipage}[t]{0.98\linewidth}
        \centering
        \includegraphics[width=\linewidth]{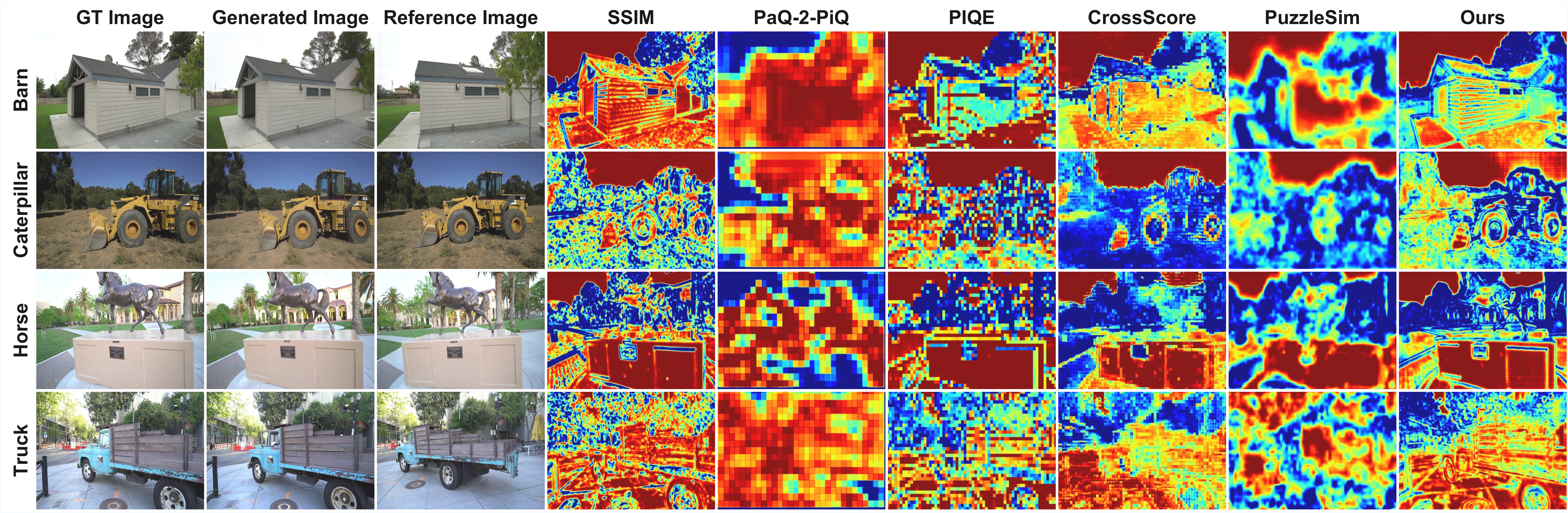}
        \captionof{figure}{Additional quality map comparisons on Tanks and Temples dataset (SSIM target). Ours$_{\text{SSIM}}$ maintains consistent quality estimation in both textured and smooth regions, demonstrating superior performance over baseline methods in complex outdoor environments.}
        \label{F8_ssim_tandt}
    \end{minipage}
    
    \vspace{5mm}
    
    \begin{minipage}[t]{0.98\linewidth}
        \centering
        \includegraphics[width=\linewidth]{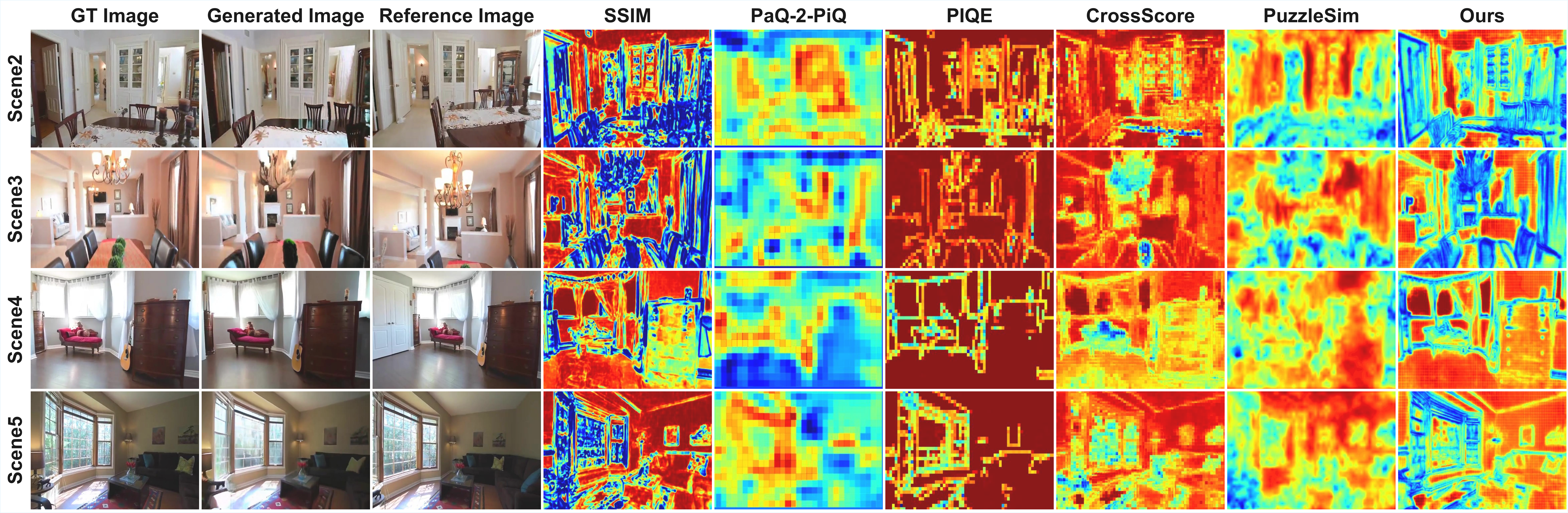}
        \captionof{figure}{Additional quality map comparisons on RealEstate10K dataset (SSIM target). Our method accurately predicts SSIM maps, producing smooth and consistent results while baselines exhibit noisy or inconsistent predictions.}
        \label{F9_ssim_re10k}
    \end{minipage}
\end{figure*}

We provide extended qualitative comparisons for SSIM-based quality assessment. Figs.~\ref{F7_ssim_mipnerf}, \ref{F8_ssim_tandt}, and \ref{F9_ssim_re10k} display results for the Mip-NeRF 360, Tanks and Temples, and RealEstate10K datasets, respectively. Notably, our $\text{Ours}_{\text{SSIM}}$ variant substantially outperforms CrossScore, despite both methods sharing the same SSIM target. This performance gap highlights the effectiveness of our reference-conditioned cross-attention and quality completion framework. Visually, $\text{Ours}_{\text{SSIM}}$ maintains consistent quality estimation across both textured and smooth regions, whereas baselines frequently exhibit noisy predictions. This validates that our framework adapts robustly to diverse quality metrics.

\subsection{More Qualitative Results for 3DGS}
\begin{figure*}
    \centering
    \includegraphics[width=0.98\linewidth]{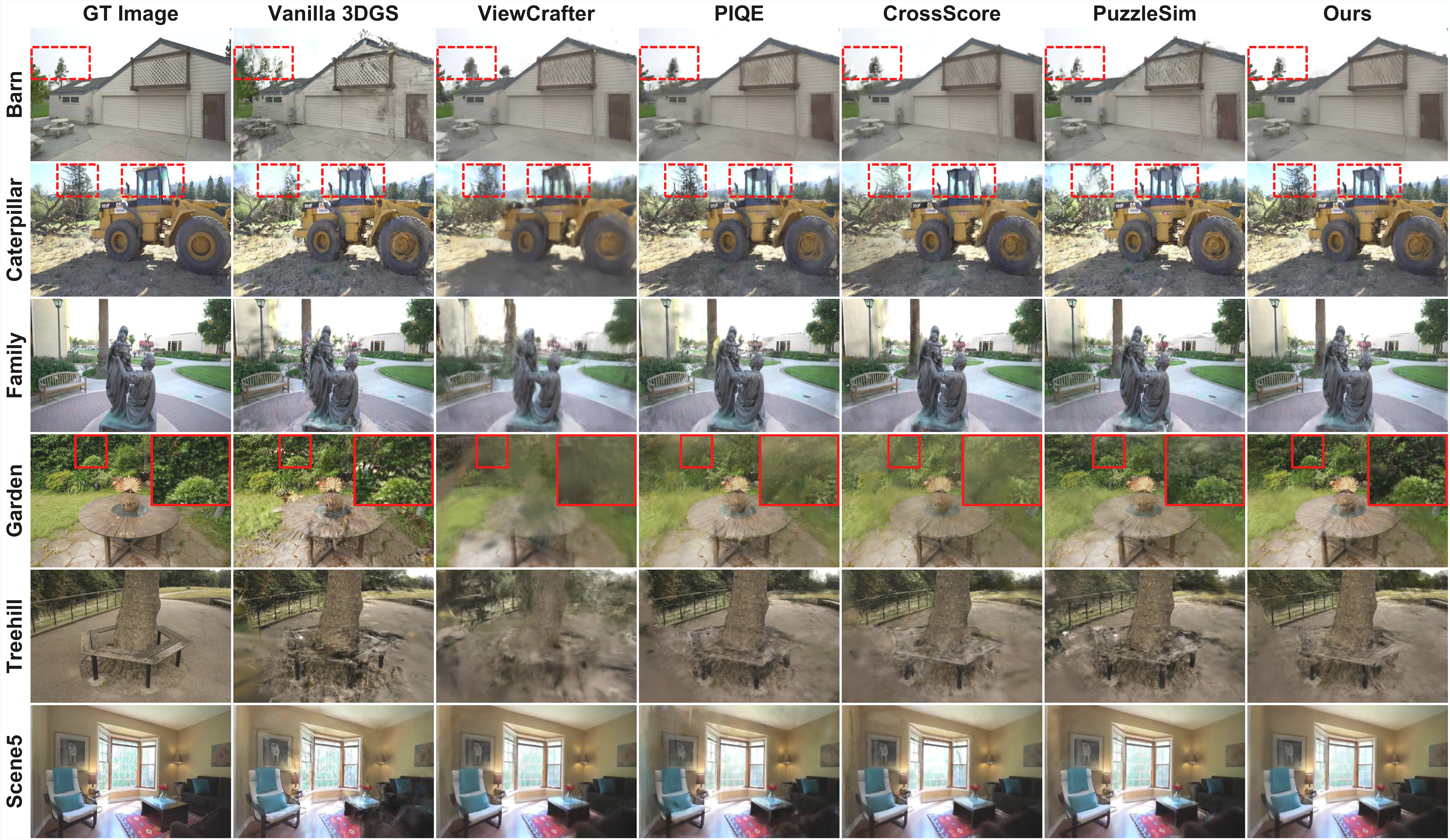}
    \vspace{-2mm}
    \caption{Qualitative comparison of 3DGS reconstruction quality. Our IQA-Guided 3DGS produces sharper geometry and more accurate textures compared to baselines by focusing computational resources on high-quality regions. Red boxes highlight representative areas where our method demonstrates superior reconstruction quality.}
    \label{F10_gs_quali}
    \vspace{-2mm}
\end{figure*}

We present additional visualization results highlighting the impact of our IQA-Guided 3DGS framework. Fig.~\ref{F10_gs_quali} shows reconstructions from the Mip-NeRF 360, Tanks and Temples, and RealEstate10K datasets, respectively. These results illustrate how our quality-guided training effectively concentrates computational resources on high-quality regions, significantly improving the overall reconstruction quality.

\section{Limitations and Discussion}
While PR-IQA achieves state-of-the-art performance in CR-IQA and significantly enhances sparse-view 3DGS reconstruction, we acknowledge several limitations and outline avenues for future research.

First, PR-IQA is currently trained using pseudo-GT quality maps derived from FR metrics, specifically DINOv2 feature similarity or SSIM. While this proxy-supervision strategy is practical for our targeted downstream task and has proven effective for geometric reconstruction, it does not fully replace human perceptual quality assessment. The model's upper bound is inherently limited by the capability of the chosen FR metric to capture perceptual subtleties or domain-specific artifacts. Incorporating human annotations or learning from large-scale perceptual preference data~\cite{kirstain2023pick} remains an exciting direction to align the quality predictions more closely with human visual perception.

Second, our experimental validation covers multiple standard benchmarks (Mip-NeRF 360, Tanks and Temples, RealEstate10K) and utilizes widely adopted backbones like ViewCrafter for view synthesis and standard 3DGS for reconstruction. However, the fields of generative AI and 3D vision are rapidly evolving, with new multi-view diffusion models and reconstruction primitives emerging frequently. A truly comprehensive evaluation across all recent architectures is beyond the scope of this work. Exploring the broader applicability of PR-IQA as a plug-and-play module for diverse generative pipelines and reconstruction methods is an interesting avenue for future research.

Finally, our framework relies on the generation of a partial quality map $\hat{Q}$, which is constructed using geometric correspondences (via VGGT and dense stereo). While our ablation studies demonstrate robustness to significant geometric noise, extreme scenarios, such as large textureless regions or severe lighting changes where stereo matching fails completely, could inevitably degrade the quality of the partial map. Future work could investigate end-to-end joint training strategies that simultaneously optimize for geometric alignment and quality estimation to mitigate this dependency.

\clearpage

\end{document}